\pdfoutput=1

\documentclass[11pt]{article}

\usepackage[final]{acl}

\usepackage{times}
\usepackage{latexsym}

\usepackage[T1]{fontenc}

\usepackage[utf8]{inputenc}
\usepackage{CJKutf8} 

\usepackage{microtype}

\usepackage{tcolorbox}
\usepackage{inconsolata}
\usepackage{subcaption}
\usepackage{graphicx}
\usepackage{xcolor}

\usepackage{booktabs, tabularx, multirow, multicol, makecell, colortbl}

\usepackage{enumitem}

\usepackage{bbding}

\usepackage{amsmath, amssymb}

\usepackage{marvosym}
\usepackage{float}
\usepackage{footmisc}
\title{MUCAR: Benchmarking Multilingual Cross-Modal Ambiguity Resolution for Multimodal Large Language Models}

\author{\textbf{Xiaolong Wang}$^{*,1,3}$, \textbf{Zhaolu Kang}$^{*,4}$, \textbf{Wangyuxuan Zhai}$^{*,6}$, \textbf{Xinyue Lou}$^{6}$, \textbf{Yunghwei Lai}$^{1}$, \\ \textbf{Ziyue Wang}$^{1}$, \textbf{Yawen Wang}$^{1}$, \textbf{Kaiyu Huang}$^{6}$, \textbf{Yile Wang}\textsuperscript{\Letter,5}, \textbf{Peng Li}\textsuperscript{\Letter,2}, \textbf{Yang Liu}$^{1,2}$\\
  \textsuperscript{1}Dept. of Comp. Sci. \& Tech., Institute for AI, Tsinghua University, Beijing, China \\
  \textsuperscript{2}Institute for AI Industry Research (AIR), Tsinghua University, Beijing, China \\
 \textsuperscript{3}Jiuquan Satellite Launch Center (JSLC), Gansu, China\\ 
 \textsuperscript{4}School of Software \& Microelectronics, Peking University, Beijing, China\\
  \textsuperscript{5}\fontsize{11pt}{10pt}\selectfont College of Computer Science and Software Engineering, Shenzhen University, Shenzhen, China\\
  \textsuperscript{6} Beijing Jiaotong University, Beijing, China\\
  \texttt{\fontsize{11pt}{10pt}\selectfont {wangxl22}@mails.tsinghua.edu.cn, kangzl9966@gmail.com, zhaiwangyuxuan@bjtu.edu.cn}\\
  \texttt{\fontsize{11pt}{10pt} \selectfont{wangyile}@szu.edu.cn, lipeng@air.tsinghua.edu.cn, liuyang2011@tsinghua.edu.cn}
  }

\DefineFNsymbols*{1}{*}
\setfnsymbol{1}

\begin{document}
\maketitle

\renewcommand{\thefootnote}{\fnsymbol{footnote}} 
    \footnotetext[1]{Equal contribution.}
\renewcommand{\thefootnote}{\arabic{footnote}}

\DefineFNsymbols*{1}{\Letter}
\setfnsymbol{1}
\renewcommand{\thefootnote}{\fnsymbol{footnote}} 
    \footnotetext[1]{Corresponding authors.}
\renewcommand{\thefootnote}{\arabic{footnote}}

\begin{abstract}
Multimodal Large Language Models (MLLMs) have demonstrated significant advances across numerous vision-language tasks. MLLMs have shown promising capability in aligning visual and textual modalities, allowing them to process image–text pairs with clear and explicit meanings. However, resolving the inherent ambiguities present in real-world language and visual contexts remains a challenge. Existing multimodal benchmarks typically overlook linguistic and visual ambiguities, relying mainly on unimodal context for disambiguation and thus failing to exploit the mutual clarification potential between modalities. To bridge this gap, we introduce MUCAR, a novel and challenging benchmark designed explicitly for evaluating multimodal ambiguity resolution across multilingual and cross-modal scenarios. MUCAR includes: (1) a multilingual dataset where ambiguous textual expressions are uniquely resolved by corresponding visual contexts, and (2) a dual-ambiguity dataset that systematically pairs ambiguous images with ambiguous textual contexts, with each combination carefully constructed to yield a single, clear interpretation through mutual disambiguation. Extensive evaluations involving 19 state-of-the-art multimodal models--encompassing both open-source and proprietary architectures--reveal substantial gaps compared to human-level performance, highlighting the need for future research into more sophisticated cross-modal ambiguity comprehension methods, further pushing the boundaries of multimodal reasoning.
\end{abstract}

\section{Introduction}
\begin{figure}[t!]
    \centering
    \includegraphics[scale=0.30]{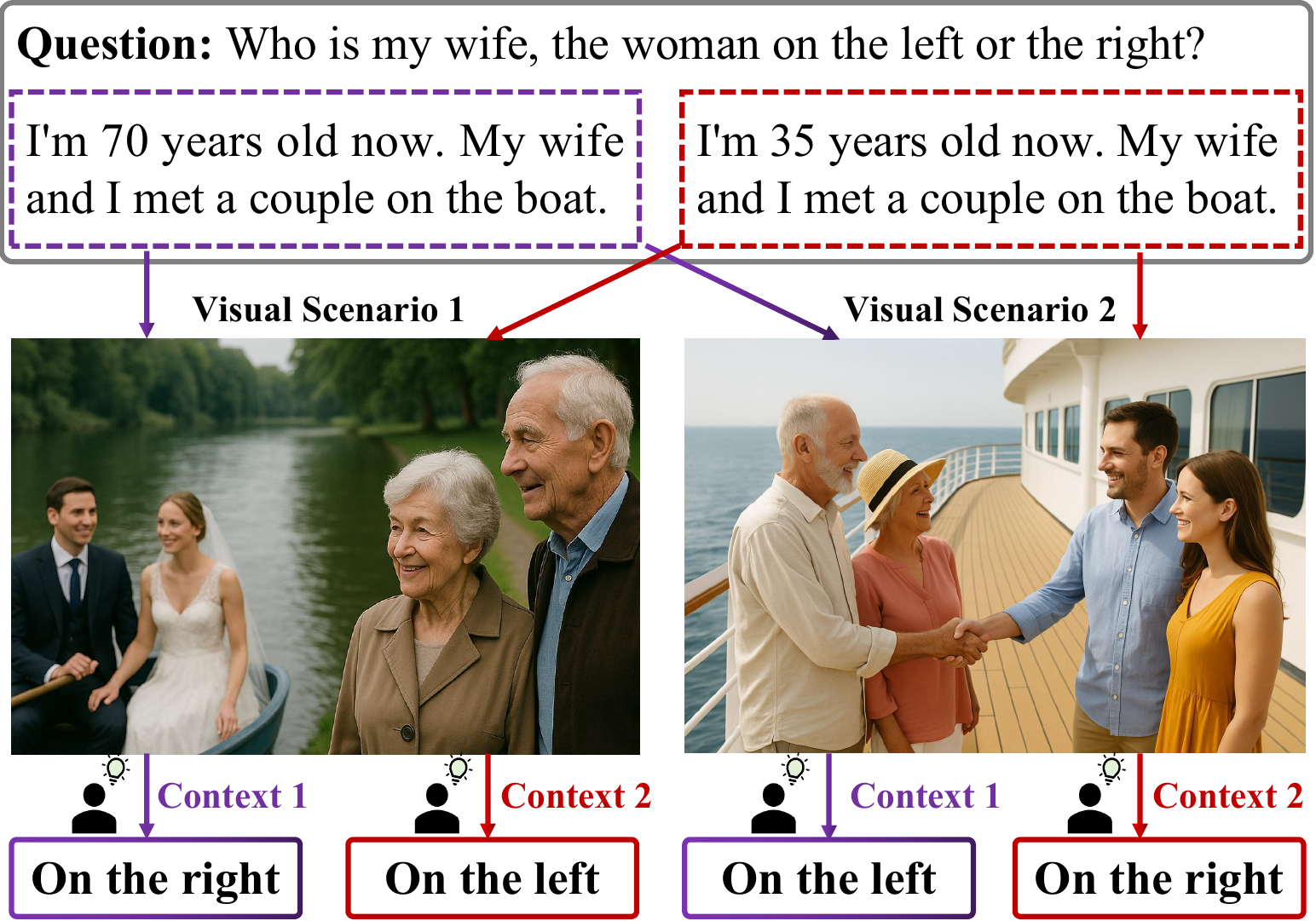}
    \caption{The interpretation of ambiguous text and visuals can be shaped by context and scenario. For instance, in “My wife and I met a couple on the boat,” it is unclear whether “on the boat” modifies “met” or “a couple,” while the image leaves the identity of the wife uncertain. Varying the visual setting (e.g., riverbank vs. cruise deck) and textual cues (e.g., age of the speaker) yields different resolutions. Each of the four context-image combinations leads to a distinct answer, with reasoning color-coded: \textcolor[RGB]{113,49,160}{purple} (Context 1), \textcolor[RGB]{208,53,44}{red} (Context 2).}

    \label{fig:introduction}
    \vspace{-1em}
\end{figure}
Multimodal Large Language Models (MLLMs; \citealt{openai2023gpt4vision, gpt-4o, liu2023improved, dai2023instructblip}) have advanced significantly in handling diverse inputs like text and images, with contextual understanding being key to their success in tasks such as question answering~\citep{shao2023prompting, liu2023cross, antol2015vqa}, image captioning~\citep{luo2023semantic, wang2023controllable, chen2015microsoft}, and multimodal reasoning~\citep{gupta2023visual, chen2023see, zellers2019recognition}. As MLLMs evolve, the ability to integrate multimodal context becomes crucial for accurate responses, underscoring the need for thorough evaluation of their contextual comprehension in real-world settings.

\begin{table*}[t]
\centering
\small
\setlength{\tabcolsep}{2pt}
\resizebox{0.98\textwidth}{!}{
\begin{tabular}{lcccc}
\toprule
\textbf{Benchmark} & \textbf{Visual Ambiguity} & \textbf{Context Ambiguity}  & \textbf{Multi-Languages}&\textbf{Evaluator} \\
\midrule
MME~\citep{fu2023mme}  & \color{purple}{\XSolidBrush} & \color{purple}{\XSolidBrush} & \color{purple}{\XSolidBrush}  & Metrics \\
MMBench~\citep{liu2023mmbench} & \color{purple}{\XSolidBrush} & \color{purple}{\XSolidBrush} &  \color{purple}{\XSolidBrush} & GPT \\
MMT-Bench~\citep{ying2024mmt}  & \color{purple}{\XSolidBrush} & \color{purple}{\XSolidBrush} &  \color{purple}{\XSolidBrush} & GPT \\
MMStar~\citep{chen2024rightwayevaluatinglarge} & \color{purple}{\XSolidBrush} & \color{purple}{\XSolidBrush} &   \color{purple}{\XSolidBrush} & Metrics \\
\midrule
HallusionBench~\citep{guan2023hallusionbench}  & \color{teal}{\Checkmark} & \color{purple}{\XSolidBrush} & \color{purple}{\XSolidBrush}  & Metrics \\
CODIS~\citep{codis}  & \color{teal}{\Checkmark} & \color{purple}{\XSolidBrush} & \color{purple}{\XSolidBrush}  & Human / GPT \\
Illusory VQA~\citep{illusoryvqa} & \color{teal}{\Checkmark} & \color{purple}{\XSolidBrush} & \color{purple}{\XSolidBrush}  & Human \\
MHaluBench~\citep{MHaluBench} & \color{teal}{\Checkmark} & \color{purple}{\XSolidBrush} & \color{purple}{\XSolidBrush}  & GPT \\
\midrule
MMA~\citep{wang2024mma}& \color{purple}{\XSolidBrush} & \color{teal}{\Checkmark} &  \color{purple}{\XSolidBrush}  & Metrics\\
VAGUE~\citep{2024arXiv241114137N} & \color{purple}{\XSolidBrush} & \color{teal}{\Checkmark} & \color{purple}{\XSolidBrush} & Metrics \\
3AM~\citep{3am-ambiguity} & \color{purple}{\XSolidBrush} & \color{teal}{\Checkmark} & \color{teal}{\Checkmark}  & Metrics \\
UNPIE~\citep{chung-etal-2024-visual} & \color{purple}{\XSolidBrush} & \color{teal}{\Checkmark} & \color{teal}{\Checkmark} & GPT \\
\midrule
\textbf{MUCAR} (Ours) & \color{teal}{\Checkmark} & \color{teal}{\Checkmark} &  \color{teal}{\Checkmark} & Human / GPT \\\bottomrule
\end{tabular}}
\caption{Comparison of our proposed \textbf{MUCAR} with recent vision-language benchmarks.}
\label{tab:other_benchmarks}
\end{table*}

Prior studies have largely emphasized tasks with clear and unambiguous inputs~\citep{fu2023mme,ying2024mmt,li2023seed}, frequently neglecting the ambiguity that naturally arises in both visual and textual modalities. Consider the example in Figure~\ref{fig:introduction}, neither the context nor the image alone can resolve the question ``Who is my wife, the woman on the left or the right?'', where a single sentence or image can often support multiple plausible interpretations depending on the specific scenario or provided context. Figure~\ref{fig:introduction} illustrates this challenge clearly. Consider the sentence ``My wife and I met a couple on the boat.'' This sentence contains structural ambiguity: it is unclear whether ``on the boat'' modifies the verb ``met'' (indicating the location of the meeting) or the noun phrase ``a couple'' (specifying the location of the couple). Simultaneously, the accompanying image introduces visual ambiguity concerning the referent of ``the wife'' among the depicted women. Notably, the ambiguity cannot be resolved independently within either modality; instead, mutual disambiguation arises when different textual scenarios (e.g., differing speaker ages) combine with different visual contexts (riverbank vs. cruise deck scenarios). Each unique combination yields a single, unambiguous interpretation, showing that textual and visual ambiguities can mutually clarify each other.

To systematically evaluate the capabilities of MLLMs to resolve such complex multimodal ambiguities, we introduce \textbf{MUCAR}, a novel benchmark specifically designed for \textbf{MU}ltilingual \textbf{C}ross-modal \textbf{A}mbiguity \textbf{R}esolution. Table~\ref{tab:other_benchmarks} summarizes recent benchmarks designed to evaluate MLLMs in terms of \textit{visual ambiguity}, \textit{contextual ambiguity}, \textit{multilinguality}, and the type of \textit{evaluator} used (e.g., metrics, GPT, or human annotations). While early benchmarks such as MMT-Bench~\citep{ying2024mmt}, MMStar~\citep{liu2023mmbench}, and MME-RealWorld~\citep{li2023seed} focus on general multimodal tasks, they lack coverage of ambiguity-related phenomena. More recent benchmarks like HallusionBench~\citep{guan2023hallusionbench}, Illusory VQA~\citep{illusoryvqa}, and CODIS~\citep{codis} begin to explore visual ambiguity, but often overlook contextual disambiguation or multilingual diversity. Notably, only a few benchmarks incorporate human evaluation, which is essential for assessing ambiguity understanding. To the best of our knowledge, MUCAR is the first benchmark to comprehensively address visual ambiguity, contextual ambiguity, and multilinguality, while integrating both human and GPT-based evaluation. This design enables a more rigorous and realistic assessment of ambiguity resolution capabilities in multimodal large language models. Our dataset and code are available at \url{https://github.com/THUNLP-MT/MUCAR}.

To summarize, our main contributions are:
\vspace{-6pt}
\begin{itemize}
    \item We construct \textbf{MUCAR}, the first \textit{multilingual cross-modal ambiguity resolution benchmark}, featuring 1278 manually curated samples in Chinese, English, and Malay, including uniquely designed dual-ambiguity cases.
    \vspace{-6pt}
    
    \item We systemically evaluate \textbf{19 sota MLLMs} (both open-source and closed-source), revealing significant limitations in resolving multilingual multimodal ambiguities.
    \vspace{-6pt}
    
    \item We propose a \textbf{simple yet effective agent-based framework} for multimodal disambiguation, which improves performance through explicit cross-modal reasoning.
\end{itemize}




\section{Related Work}
\begin{figure*}[h!]
    \centering
    \includegraphics[width=1\linewidth]{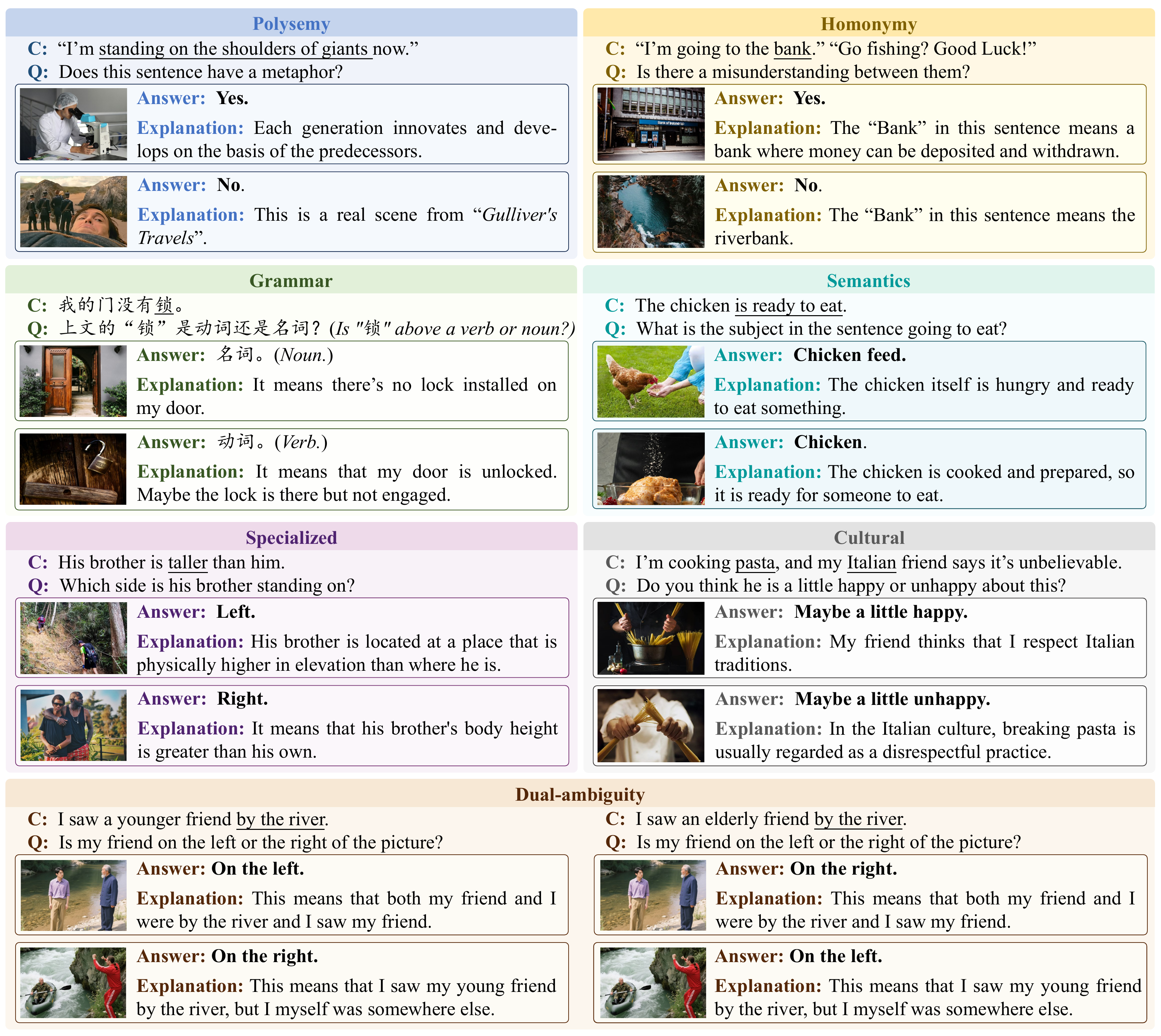}
    \vspace{-12pt}
    \caption{Taxonomy of our benchmark. We present one example for each category. Each example includes a context (C), a question (Q), and two different images with their corresponding answers and explanations.}
    \label{fig:introduction}
    \vspace{-6pt}
\end{figure*}

\textbf{Context Ambiguity} Ambiguity is an inherent characteristic of linguistic text, emerging naturally due to the potential for multiple interpretations, especially in open-domain question answering tasks~\citep{ambigqa,IterativePrompting}. Existing research on ambiguity resolution in language primarily follows two directions. One line leverages contextual cues from surrounding text to resolve ambiguity, as seen in ~\citet{ambiguityaware, agenticveriamb}, which uses in-context learning to disambiguate textual inputs. Another line exploits visual information to disambiguate language, particularly for polysemous word translation, such as in 3AM~\citep{3am-ambiguity}, which aligns ambiguous words with visual semantics. 

Aligned with the multimodal ambiguity setting in MMA~\citep{wang2024mma}, our work focuses on using visual input to disambiguate multilingual expressions. Beyond standard NLP ambiguities, we also address those arising from domain-specific and cultural differences. In particular, we construct a benchmark featuring both textual and visual ambiguity across multiple languages, aiming to evaluate the ability of MLLMs to resolve complex multimodal ambiguities in realistic scenarios.

\textbf{Visual Ambiguity} Visual ambiguity often stems from incomplete visual cues or interfering noise in the scene~\citep{denison2018humans}. Most previous vision-language benchmarks assume unambiguous~\citep{liu2023mmbench,fu2023mme,liu2023mmbench,li2023seed} input or highlight the visual ambiguities caused by optical illusions~\citep{guan2023hallusionbench, illusoryvqa, cui2023holistic, fu2023challenger}. Early Multimodal datasets like MS-COCO \citep{chen2015microsoft} focus on literal descriptions, while later works, e.g., CODIS~\citep{codis} highlight the need for diverse context to reflect multiple valid interpretations.
Inspired by CODIS ~\citep{codis}, we assess the capability of MLLMs to disambiguate visual ambiguity through textual modalities instead of just recognizing ambiguities. Different from CODIS, we construct challenging dual-ambiguity instances, combining ambiguous visuals and texts that jointly resolve into a single interpretation, further testing the limits of multimodal reasoning.

\section{MUCAR}


MUCAR is proposed for evaluating the capabilities of MLLMs in image-dependent context disambiguation. Figure \ref{fig:introduction} presents several examples from our benchmark, highlighting the diversity of contexts covered. In this section, we first describe our taxonomy of context. Then, we delve into the instruction design. Finally, we introduce data collection procedures.

\subsection{Taxonomy}

Given the extensive and varied nature of context information, comprehensive cataloging of all forms of context is challenging. With the aim of establishing an outstanding benchmark for disambiguation, we identified seven representative types. The first six types are inspired by the information people require to understand context. When collecting the data, we surprisingly found that in some cases, the combination of ambiguous text and ambiguous images led to mutual disambiguation. In other words, neither modality alone provided sufficient clarity, but together they resolved the ambiguity inherent in both. This observation inspired us to define the seventh type and collect the corresponding data. \textit{To the best of our knowledge, we are the first to construct data for this type.} Figure~\ref{fig:introduction} illustrates examples with corresponding classification explanations.


\noindent \textbf{Polysemy.} \citet{HomonymyAndPolysemy} provides one formalization of polysemy, by referring to a word with two or more related meanings. These related meanings often share a conceptual core, with one meaning typically being an extension or variation of the other. Some contexts can be interpreted both literally and metaphorically. The ``Polysemy'' part in Figure \ref{fig:introduction} gives a good example.

\noindent \textbf{Homonymy.} \citet{HomonymyAndPolysemy} also provides one formalization of homonymy. Opposite to polysemy, homonymy refers to a word having two or more unrelated meanings that stem from different historical origins, and the meanings of homonyms have no inherent connection. Disambiguating homonyms relies on the other elements. The ``Homonymy'' part in Figure \ref{fig:introduction} serves as a good example.

\begin{CJK*}{UTF8}{gbsn}

\noindent \textbf{Grammar.} This ambiguity occurs when sentence structures allow for multiple interpretations, often due to the placement of words or phrases. Such structural issues can make it unclear which part of the sentence a modifier applies to or the relationship between different clauses. The ``Grammar'' part in Figure \ref{fig:introduction} provides a clear illustration. As the ``Grammar'' part in the figure shows, in Chinese, ``我的门没有锁'' can be interpreted as ``My door does not have a lock'' or ``My door has not been locked'', here ``锁'' can be understood as a noun or verb, which leads to ambiguity.

\end{CJK*}

\noindent \textbf{Semantics.} Understanding the timing and sequence of events is crucial when we understand a context. However, an isolated context can only provide us with static information, which is insufficient for dynamic events. The disambiguation can only be achieved when unambiguous image gives us more information. The ``Semantics'' case in Figure~\ref{fig:introduction} provides a representative example.


\noindent \textbf{Specialized.}  We define Specialized taxonomy to encompass terms or concepts that have distinct meanings across academic domains or personal situations. These terms often lead to ambiguity when encountered by individuals with varying background knowledge. The ``Specialized'' part in Figure \ref{fig:introduction} presents a good example.

\noindent \textbf{Cultural.} Some context can be interpreted differently depending on the cultural background of the interpreter, as words, symbols, and actions with specific meanings and connotations. Cultural norms, values, and historical experiences shape how individuals understand and react to information. This can lead to significant textual ambiguities where meaning is lost or distorted. The ``Cultural'' part in Figure \ref{fig:introduction} offers a good example.



\noindent \textbf{Dual-ambiguity.} This data type highlights our unique contribution. In this type, the context and the image are both ambiguous, but their combined information allows for clear disambiguation. Figure \ref{fig:introduction} gives a good example. The context ambiguity covers the six former types. Regarding image ambiguity, following CODIS \citep{codis}, image ambiguity can be further categorized into distinct types, such as location and orientation, temporal information, cultural background, attributes, and relationships. However, for the purpose of this paper, we group these various types under the general term ``image ambiguity''. Thus, dual-ambiguity specifically denotes the situation where both the context and the image exhibit ambiguity.

\subsection{Instruction Design}




In order to ensure that model fully understands the context and image instead of making choices randomly, we organize our dataset in pairs. For the first six data types, the query can be represented as $(\mathcal{C}, \mathcal{Q}, \mathcal{I}_i)$. Each pair consists of an identical ambiguous context $\mathcal{C}$ and a question $\mathcal{Q}$, which are presented alongside $i$ different unambiguous images $(\mathcal{I}_1, \mathcal{I}_2, \dots, \mathcal{I}_i)$.

For the dual-ambiguity data type, we manually group queries that look similar. Even within a pair, the context $\mathcal{C}_i$, the question $\mathcal{Q}_i$, and the image $\mathcal{I}_i$ may differ for each query instance. More formally, each pair can be represented as $(\mathcal{C}_i, \mathcal{Q}_i, \mathcal{I}_i)$.

Each $(\mathcal{C}, \mathcal{Q}, \mathcal{I})$ is independently input into MLLMs as a query without being influenced by other queries in the same pair. MLLMs receive $i$ queries in the same pair independently and produce their outputs $(\mathcal{O}_1, \mathcal{O}_2, \dots, \mathcal{O}_i)$. These outputs are evaluated by comparing them with ground truth answers $(\mathcal{A}_1, \mathcal{A}_2, \dots, \mathcal{A}_i)$.

\begin{figure}
    \centering
    \includegraphics[width=0.96\linewidth]{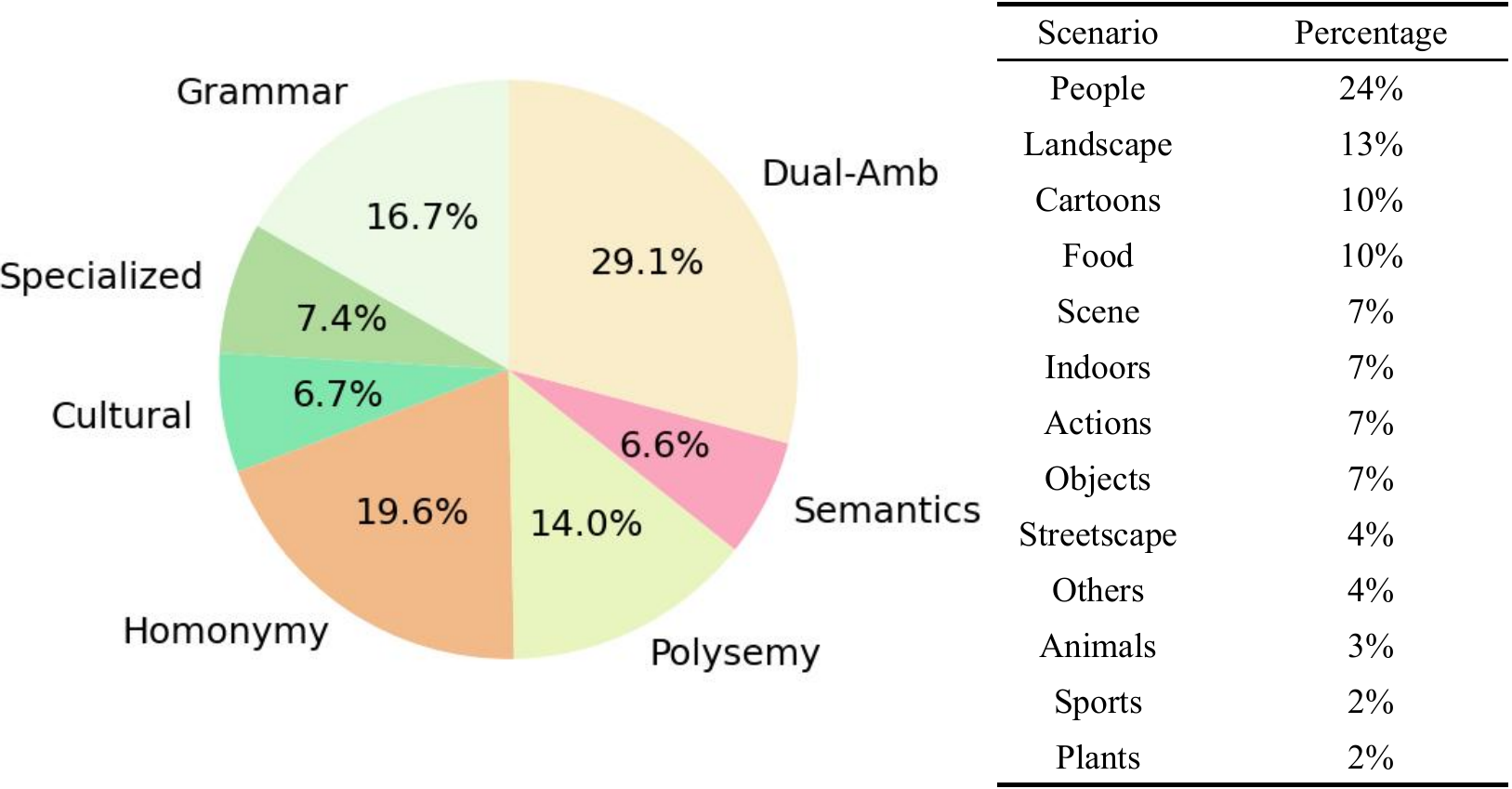}
    \caption{Distribution of seven categories (left) and scenarios (right) of our benchmark.}
    \label{fig:dis}
\end{figure}

\subsection{Data Collection}

In this section, we detail how we construct this benchmark by using a three-step collection process.

\noindent \textbf{Context Collection.} We manually collected ambiguous contexts which can only be resolved with external images. Most of the contexts come from the Internet, while others are created manually. These contexts span three languages: Chinese, English, and Malay. We meticulously reviewed every context to exclude any that were unambiguous. In total, we collected 1278 contexts. The first six types comprise 906 contexts, and the remaining 372 contexts belong to the dual-ambiguity type.

\noindent \textbf{Design of Questions, Images and Answers.} For each context, we manually wrote questions and answers. As for images, we also manually collected them from the Internet or designed them ourselves. Specifically, for the dual-ambiguity type, the majority of images are sourced from CODIS \citep{codis}, a benchmark constructed using ambiguous images. The data are compiled with following rules:




(1) Questions are designed to target ambiguous aspects within the contexts. Disambiguation of these contexts is not possible without the inclusion of external images.

(2) For each context and question, every unique image associated with them should lead to a distinct interpretation of the context, resulting in a unique answer per image. Crucially, the answer cannot be determined from the image or the context in isolation. The answer can be determined only when we give MLLMs the query in the format $(\mathcal{C}, \mathcal{Q}, \mathcal{I})$.

(3) To balance the performance and evaluation efficiency, MLLMs were required to generate outputs following a specific template with clear, predefined options. This method ensures the objectivity of responses and facilitates their efficient evaluation.

\noindent \textbf{Data Verification.} Five annotators participated in this process. To ensure dataset quality, each submission was cross-checked by the remaining four annotators. Data were retained only if they satisfied the following conditions: (1) correctness, (2) distinctiveness from existing data, and (3) compliance with all predefined criteria. Submissions that did not meet these conditions were returned to the annotator for revision.


Finally, our benchmark comprises 1278 queries and 501 $(\mathcal{C}, \mathcal{Q}, \mathcal{I})$ pairs, categorized into seven types. Figure \ref{fig:dis} visually represents how the categories and scenarios are distributed.

\subsection{Evaluation Metrics}




\begin{table*}[t]
\centering
\small
\vspace{-2pt}
\begin{subtable}{\textwidth}
\centering
\caption*{(a) Without confusing options}
\resizebox{\textwidth}{!}{
\begin{tabular}{l|cccccccccccccc|cc}
\toprule
\multirow{2}{*}{\textbf{Model}} & \multicolumn{2}{c}{\textbf{Polysemy}} & \multicolumn{2}{c}{\textbf{Homonymy}} & \multicolumn{2}{c}{\textbf{Grammar}} & \multicolumn{2}{c}{\textbf{Semantics}} & \multicolumn{2}{c}{\textbf{Specialized}} & \multicolumn{2}{c}{\textbf{Cultural}} & \multicolumn{2}{c|}{\textbf{Dual-ambiguity}} & \multicolumn{2}{c}{\textbf{Overall}} \\
& \textbf{Acc\textsubscript{\textit{p}}} & \textbf{Acc\textsubscript{\textit{q}}} & \textbf{Acc\textsubscript{\textit{p}}} & \textbf{Acc\textsubscript{\textit{q}}} & \textbf{Acc\textsubscript{\textit{p}}} & \textbf{Acc\textsubscript{\textit{q}}} & \textbf{Acc\textsubscript{\textit{p}}} & \textbf{Acc\textsubscript{\textit{q}}} & \textbf{Acc\textsubscript{\textit{p}}} & \textbf{Acc\textsubscript{\textit{q}}} & \textbf{Acc\textsubscript{\textit{p}}} & \textbf{Acc\textsubscript{\textit{q}}} & \textbf{Acc\textsubscript{\textit{p}}} & \textbf{Acc\textsubscript{\textit{q}}} &
\textbf{Acc\textsubscript{\textit{p}}} & \textbf{Acc\textsubscript{\textit{q}}} \\
\midrule
\rowcolor{gray!20}\multicolumn{17}{c}{\textbf{API-based Models}} \\\midrule

GPT-4V &  35.23 & 55.87 & 32.48 & 53.20 & \textbf{43.93} & \textbf{65.89} & 37.21 & 60.71 & 29.55 & 60.00 & 19.05 & 38.82 & 20.62 & 42.74 & 30.39 & 52.70\\
GPT-4o & \textbf{36.36} & \textbf{61.45} & \textbf{44.44} & \textbf{64.00} & 41.12 & 65.42 & \textbf{44.19} & \textbf{64.29} & \textbf{47.73} & \textbf{70.53} & \textbf{28.57} & \textbf{47.06} & 21.65 & \textbf{53.23} & \textbf{34.96} & \textbf{60.13} \\
Gemini-2.0-flash &  19.48 & 52.87 & 38.18 & 63.14 & 17.78 & 50.56 & 19.51 & 50.00 & 30.77 & 63.53 & 25.00 & 46.91 & 12.36 & 36.16 & 23.46 & 52.11\\
Claude-3.5-Sonnet & 3.41 & 16.76 & 2.56 & 16.00 & 5.61 & 21.03 & 0.00 & 14.29 & 6.82 & 17.89 & 0.00 & 5.88 & \textbf{23.28} & 51.10 & 9.37 & 26.36\\

\midrule
\rowcolor{gray!20}\multicolumn{17}{c}{\textbf{Open-source Models $> 7\mathrm{B}$}} \\\midrule

Kimi-VL & 3.37 & 36.31 & 5.13 & 46.40 & 5.56 & 43.46 & 2.38 & 40.48 & 2.27 & 36.84 & 0.00 & 47.06 & 14.08 & 52.27 & 4.74 & 45.15 \\
Llama-3.2-Vision-11B & 33.71 & 63.69 & 37.61 & 66.80 & 25.93 & 59.35 & 28.57 & 57.14 & 27.27 & 63.16 & 19.51 & 58.82 & 14.08 & 52.27 & 28.09 & 59.45 \\
MiniCPM-o 2.6 & 38.20 & 66.48 & 38.46 & 66.40 & \textbf{33.33} & 63.55 & 28.57 & 60.71 & \textbf{40.91} & \textbf{69.47} & \textbf{55.29} & 29.27 & 4.23 & 20.53 & 34.00 & 53.10 \\
Idefics3-8B-Llama3 & 39.33 & 67.04 & 37.31 & 62.80 & 31.48 & 62.62 & 35.71 & 65.48 & 31.32 & 56.84 & 24.09 & 56.47 & 19.72 & 49.87 & 32.38 & 58.91 \\
InternVL2-8B & 20.24 & 49.56 & 23.76 & 53.67 & 10.29 & 40.44 & 23.81 & 47.62 & 20.00 & 47.13 & 21.95 & 48.24 & 20.97 & 47.58 & 20.08 & 48.16 \\
InternVL2.5-8B-MPO & 42.26 & 69.32 & 43.56 & 68.81 & 25.00 & 60.29 & \textbf{42.86} & \textbf{69.05} & 37.50 & 58.62 & 19.51 & 57.65 & 24.19 & 51.88 & 35.73 & 61.68 \\
InternVL2.5-8B-MPO-AWQ & \textbf{42.86} & \textbf{69.91} & \textbf{46.53} & \textbf{70.18} & 30.88 & \textbf{64.71} & 33.33 & 61.90 & 37.50 & 60.92 & 21.95 & \textbf{60.00} & \textbf{25.81} & \textbf{54.30} & \textbf{37.32} & \textbf{63.32} \\

\midrule
\rowcolor{gray!20}\multicolumn{17}{c}{\textbf{Open-source Models $\leq 7\mathrm{B}$}} \\\midrule

Deepseek-VL-Tiny & 0.00 & 44.69 & 5.98 & 51.20 & 0.00 & 48.60 & 2.33 & 48.81 & 6.82 & 47.37 & 0.00 & 49.41 & 16.49 & \textbf{54.03} & 6.77 & 50.12 \\
LLaVA-v1.6-vicuna-7b & 15.73 & 54.19 & 21.37 & 54.80 & 14.81 & 54.21 & 21.43 & 55.95 & 11.36 & 42.11 & 12.20 & 51.76 & 16.90 & 50.40 & 16.46 & 52.26 \\
LLaVA-v1.6-mistral-7b & 19.11 & 55.87 & 23.08 & 58.00 & 12.96 & 54.67 & 16.67 & 58.33 & 15.91 & 49.47 & 2.44 & 47.06 & 16.90 & 50.40 & 16.26 & 53.60 \\
Qwen2.5-VL-3B-Instruct & 33.93 & 64.60 & 27.72 & 56.88 & 17.65 & 53.68 & 19.05 & 50.00 & 22.50 & 56.32 & 12.20 & 55.29 & 20.97 & 40.86 & 25.55 & 53.56 \\
Qwen2.5-VL-7B-Instruct & \textbf{34.52} & 60.47 & 32.67 & 60.09 & 29.41 & 58.09 & \textbf{33.33} & 50.00 & \textbf{35.00} & \textbf{63.22} & 14.63 & 54.12 & 19.35 & 49.19 & \textbf{29.94} & 56.29 \\
mPLUG-Owl3-7B-240728 & 32.74 & 63.42 & 35.64 & 63.30 & 22.06 & 58.82 & 28.57 & 61.90 & 22.50 & 55.17 & \textbf{48.24} & 14.63 & \textbf{27.42} & 52.96 & 28.74 & 58.25 \\
mPLUG-Owl3-2B-241014 & 33.73 & \textbf{65.68} & 36.08 & 64.71 & 27.47 & \textbf{62.09} & 30.95 & \textbf{62.20} & 30.00 & 60.92 & 10.00 & 55.74 & 16.49 & 51.61 & 25.65 & \textbf{59.29} \\
LLaVA-OneVision & 30.33 & 61.45 & \textbf{38.46} & \textbf{64.80} & \textbf{30.56} & 61.22 & 28.57 & 58.33 & 27.27 & 55.79 & 21.95 & \textbf{60.00} & 9.86 & 47.20 & 28.37 & 57.20 \\

\bottomrule
\end{tabular}
}
\end{subtable}

\vspace{-3pt}

\begin{subtable}{\textwidth}
\centering
\caption*{(b) With confusing options}
\resizebox{\textwidth}{!}{
\begin{tabular}{l|cccccccccccccc|cc}
\toprule
\multirow{2}{*}{\textbf{Model}} & \multicolumn{2}{c}{\textbf{Polysemy}} & \multicolumn{2}{c}{\textbf{Homonymy}} & \multicolumn{2}{c}{\textbf{Grammar}} & \multicolumn{2}{c}{\textbf{Semantics}} & \multicolumn{2}{c}{\textbf{Specialized}} & \multicolumn{2}{c}{\textbf{Cultural}} & \multicolumn{2}{c|}{\textbf{Dual-ambiguity}} & \multicolumn{2}{c}{\textbf{Overall}} \\
& \textbf{Acc\textsubscript{\textit{p}}} & \textbf{Acc\textsubscript{\textit{q}}} & \textbf{Acc\textsubscript{\textit{p}}} & \textbf{Acc\textsubscript{\textit{q}}} & \textbf{Acc\textsubscript{\textit{p}}} & \textbf{Acc\textsubscript{\textit{q}}} & \textbf{Acc\textsubscript{\textit{p}}} & \textbf{Acc\textsubscript{\textit{q}}} & \textbf{Acc\textsubscript{\textit{p}}} & \textbf{Acc\textsubscript{\textit{q}}} & \textbf{Acc\textsubscript{\textit{p}}} & \textbf{Acc\textsubscript{\textit{q}}} & \textbf{Acc\textsubscript{\textit{p}}} & \textbf{Acc\textsubscript{\textit{q}}} &
\textbf{Acc\textsubscript{\textit{p}}} & \textbf{Acc\textsubscript{\textit{q}}} \\
\midrule
\rowcolor{gray!20}\multicolumn{17}{c}{\textbf{API-based Models}} \\\midrule

GPT-4V & 26.51 & 47.93 & 20.62 & 41.18 & 26.37 & 52.20 & 21.43 & 45.12 & 20.00 & 47.13 & 6.67 & 26.23 & 15.46 & 30.65 & 19.93 & 40.45 \\
GPT-4o & \textbf{32.53} & \textbf{50.30} & \textbf{25.77} & \textbf{45.10} & \textbf{31.87} & \textbf{56.04} & \textbf{23.81} & \textbf{46.34} & \textbf{22.50} & \textbf{51.72} & \textbf{13.33} & 29.51 & \textbf{17.53} & \textbf{36.29} & \textbf{23.92} & \textbf{44.51} \\
Gemini-2.0-flash & 8.06 & 36.80 & 8.82 & 32.41 & 14.06 & 44.53 & 8.70 & 40.91 & 4.76 & 25.58 & 6.67 & \textbf{32.26} & 11.02 & 31.23 & 10.00 & 34.85 \\
Claude-3.5-Sonnet & 3.61 & 22.49 & 4.12 & 22.55 & 4.40 & 23.08 & 2.38 & 20.73 & 2.50 & 27.59 & 3.33 & 13.11 & 15.59 & 35.64 & 7.56 & 26.50 \\

\midrule
\rowcolor{gray!20}\multicolumn{17}{c}{\textbf{Open-source Models $> 7\mathrm{B}$}} \\\midrule

Kimi-VL & 1.13 & 31.84 & 1.71 & 39.60 & 1.85 & 40.19 & 0.00 & 36.90 & 0.00 & 32.63 & 0.00 & 41.18 & 5.71 & 18.82  & 1.50 & 31.98 \\
Llama-3.2-Vision-11B & 21.35 & 56.98 & 37.61 & \textbf{63.60} & 13.89 & 43.93 & 14.29 & 45.24 & 11.37 & 48.42 & 17.07 & \textbf{54.12} & 5.71 & 34.67 & 19.87 & 48.00 \\
MiniCPM-o 2.6 & 20.22 & 53.07 & 34.18 & 60.40 & 28.70 & 58.41 & 30.95 & 59.52 & 18.18 & 60.00 & 14.63 & 50.58 & 2.86 & 12.63 & 23.50 & 44.40 \\
Idefics3-8B-Llama3 & 29.21 & 58.10 & \textbf{39.32} & 62.80 & 24.07 & 55.14 & \textbf{35.71} & \textbf{60.71} & \textbf{31.82} & \textbf{62.11} & 17.07 & 51.76 & 12.86 & \textbf{50.54} & 28.14 & 56.37 \\
InternVL2-8B & 13.48 & 35.20 & 8.55 & 33.20 & 7.41 & 34.58 & 7.14 & 36.90 & 20.45 & 41.05 & 7.32 & 36.47 & 8.06 & 31.45 & 9.78 & 34.24 \\
InternVL2.5-8B-MPO & 26.97 & 58.10 & 27.35 & 54.80 & 27.78 & 56.54 & 28.57 & 58.33 & 25.00 & 52.63 & \textbf{19.51} & 51.76 & 20.97 & 46.51 & 25.95 & 53.01 \\
InternVL2.5-8B-MPO-AWQ & \textbf{31.46} & \textbf{63.69} & 33.33 & 59.20 & \textbf{34.26} & \textbf{62.62} & 30.95 & 59.52 & 31.81 & 57.89 & 14.63 & 52.94 & \textbf{22.58} & 50.27 & \textbf{29.94} & \textbf{57.31} \\

\midrule
\rowcolor{gray!20}\multicolumn{17}{c}{\textbf{Open-source Models $\leq 7\mathrm{B}$}} \\\midrule
Deepseek-VL-Tiny & 0.00 & 21.30 & 0.00 & 21.08 & 0.00 & 30.77 & 0.00 & 24.39 & 5.00 & 31.03 & 0.00 & 27.87 & 9.28 & 35.48 & 3.47 & 28.61 \\
LLaVA-v1.6-vicuna-7b & 13.48 & 51.95 & 19.66 & 58.00 & 14.81 & 56.07 & 19.05 & 54.76 & 11.36 & 49.47 & 14.63 & \textbf{55.29} & 11.43 & 46.77 & 15.19 & 52.54 \\
LLaVA-v1.6-mistral-7b & 19.10 & 57.54 & 23.93 & 59.20 & 12.96 & 54.21 & 26.19 & \textbf{69.91} & 18.18 & 51.58 & 2.44 & 49.41 & 14.29 & 47.58 & 17.14 & 53.71 \\
Qwen2.5-VL-3B-Instruct & 25.84 & 56.42 & 28.21 & 58.00 & 25.00 & 57.94 & 19.05 & 53.57 & 11.36 & 50.53 & 14.63 & 42.35 & \textbf{24.19} & 45.43 & 22.95 & 52.23 \\
Qwen2.5-VL-7B-Instruct & 13.48 & 42.46 & 19.66 & 50.80 & 27.78 & 55.61 & 26.19 & 51.19 & 27.27 & 60.00 & 9.76 & 44.71 & 20.97 & \textbf{48.66} & 20.76 & 50.12 \\
mPLUG-Owl3-7B-240728 & 21.35 & 56.42 & 30.77 & 60.00 & 22.22 & 56.54 & 26.19 & 59.52 & 18.18 & 53.68 & 9.76 & 43.53 & \textbf{24.19} & 47.85 & 23.35 & 53.79 \\
mPLUG-Owl3-2B-241014 & 24.10 & 53.85 & 30.93 & 60.78 & 25.27 & 59.34 & \textbf{33.33} & 58.54 & \textbf{32.50} & \textbf{60.92} & \textbf{16.67} & 49.18 & 13.40 & 46.77 & 22.70 & 54.28 \\
LLaVA-OneVision & \textbf{40.44} & \textbf{68.16} & \textbf{39.32} & \textbf{65.60} & \textbf{34.26} & \textbf{63.08} & \textbf{33.33} & 64.29 & 31.82 & 60.00 & \textbf{17.07} & \textbf{55.29} & 8.57 & 41.67 & \textbf{31.40} & \textbf{57.39} \\

\midrule
\rowcolor{gray!20}\multicolumn{17}{c}{\textbf{Human}}\\\midrule
Human & 75.00 & 86.59 & 73.50 & 86.00 & 70.09 & 85.05 & 78.57 & 89.29 & 79.55 & 89.36 & 68.29 & 83.53 & 81.43 & 87.12 & 74.66 & 86.42 \\

\bottomrule
\end{tabular}
}
\end{subtable}

\caption{Results of MLLMs on MUCAR benchmark under two settings: (top) without confusing options and (bottom) with confusing options. For humans, the two experimental settings show little difference, so we chose the more challenging \textbf{with confusing options} setting for human evaluation.}
\label{tab:combined-results}
\vspace{-6pt}
\end{table*}

For the $k$-th pair of queries, we decide to use $(\mathcal{O}_{k1}, \mathcal{O}_{k2}, \dots, \mathcal{O}_{ki})$ to represent the model's outputs of a pair, and $(\mathcal{A}_{k1}, \mathcal{A}_{k2}, \dots, \mathcal{A}_{ki})$ for the corresponding ground truth answers. We express the evaluation of these model outputs as follows:
\[
\text{Eval}(\mathcal{O}_{ki})=\begin{cases}
1 & \text{if } \mathcal{O}_{ki} \text{ matches } \mathcal{A}_{ki} \\
0 & \text{otherwise}
\end{cases},  i \in \mathbb{Z}^+
\]

Following \citet{fu2023mme}, our evaluation utilizes two metrics, pair-wise accuracy $\mathrm{Acc_p}$ and query-wise accuracy $\mathrm{Acc_q}$, these metrics can be calculated as follows:
\[
\text{Acc}_p = \frac{1}{n_p} \sum_{k = 1}^{n_p} \prod_{i = 1}^{ n_k } \text{Eval}(\mathcal{O}_{ki}),
\]
\[
\text{Acc}_q = \frac{1}{n_q} \sum_{k = 1}^{n_p} \sum_{i = 1}^{ n_k } \text{Eval}(\mathcal{O}_{ki}).
\]

where $n_k$ denotes the number of queries within each pair, $n_p$ represents the total number of pairs, and $n_q$ is the total number of individual queries. $\mathrm{Acc_p}$ denotes the accuracy of judging each individual query's correctness independently. $\mathrm{Acc_q}$ requires that a pair is considered correct only if MLLMs correctly judge all queries within the pair.

\section{Experiments}

\subsection{Models}

We evaluate a total of 19 models covering a range of scales and architectures. Our evaluated proprietary models include GPT-4V~\citep{openai2023gpt4vision}, GPT-4o~\citep{gpt-4o}, Gemini~\citep{gemini2023gemini}, and Claude-3.5-Sonnet~\citep{Claude35Sonnet}. For open-source models, we include Deepseek-VL-Tiny~\citep{lu2024deepseekvl}, Kimi-VL~\citep{kimiteam2025kimivltechnicalreport}, Llama-3.2-Vision-11B~\citep{Llama32Vision11B}, MiniCPM-o 2.6~\citep{yao2024minicpm}, InternVL2.5 series~\citep{InternVL2.5-8B-MPO_wang2024mpo,InternVL2.5-8B-MPO-AWQ_chen2024expanding}, LLaVA-v1.6-vicuna-7b~\citep{LLaVA_liu2024improvedbaselinesvisualinstruction}, Qwen2.5-VL series~\citep{qwen2.5-VL}. Details of these models are listed in Table~\ref{tab:models} in Appendix~\ref{app:models}.

\subsection{Main Results}

Main experimental results on our benchmark of all 19 models are reported in Table~\ref{tab:combined-results}.

\noindent \textbf{Overall Performance.}
Across all evaluated models, InternVL2.5-8B-MPO-AWQ achieves the best overall accuracy (\textit{Acc$_q$} = 63.32\%), followed closely by InternVL2.5-8B-MPO (61.68\%) and MiniCPM-o 2.6 (59.45\%). Among proprietary models, GPT-4o outperforms the others, obtaining an overall accuracy of 60.13\%, slightly higher than GPT-4V (52.70\%). In contrast, Claude-3.5-Sonnet and Kimi-VL underperform, showing limited ability in disambiguation tasks.

\noindent \textbf{Results with Different Model Size.}
Open-source models with scales \textit{larger than 7B} generally outperform smaller ones, with all top-performing models falling within this range of scale, which is likely to benefit from richer training data and more advanced architectures. In comparison, models \textit{smaller than or equal to 7B} show a clear performance gap. Although certain models, such as LaVA-Onevision (57.20\%) and mPLUG-Owl2-2B (59.29\%), perform competitively, most smaller models struggle with complex ambiguities, particularly in semantic and cultural contexts.

\noindent \textbf{Results on Different Categories.}
We further break down the results by disambiguation categories, and find that InternVL2.5-8B-MPO-AWQ consistently leads in most categories, especially in Homonymy (70.18\%), Grammar (64.71\%), and Semantics (69.05\%). Notably, MiniCPM-o 2.6 excels in the Specialized category (69.47\%), suggesting domain knowledge plays a key role. In the Cultural category, which requires understanding cross-cultural references, models like GPT-4o (47.06\%) and InternVL2.5-8B-MPO-AWQ (60.00\%) show relatively stronger performance. On the other hand, most models perform poorly in the Polysemy and Dual-Ambiguity categories, reflecting the inherent challenges in resolving subtle or cross-modal ambiguities.

In summary, model size and architecture significantly affect cross-modal disambiguation performance. Larger models and instruction-finetuned models usually demonstrate better generalization. The disparity across categories reveals the diverse challenges in context-dependent reasoning, particularly in categories involving semantic, cultural, or compound ambiguities.

\begin{figure}
    \centering
    \includegraphics[width=0.98\linewidth]{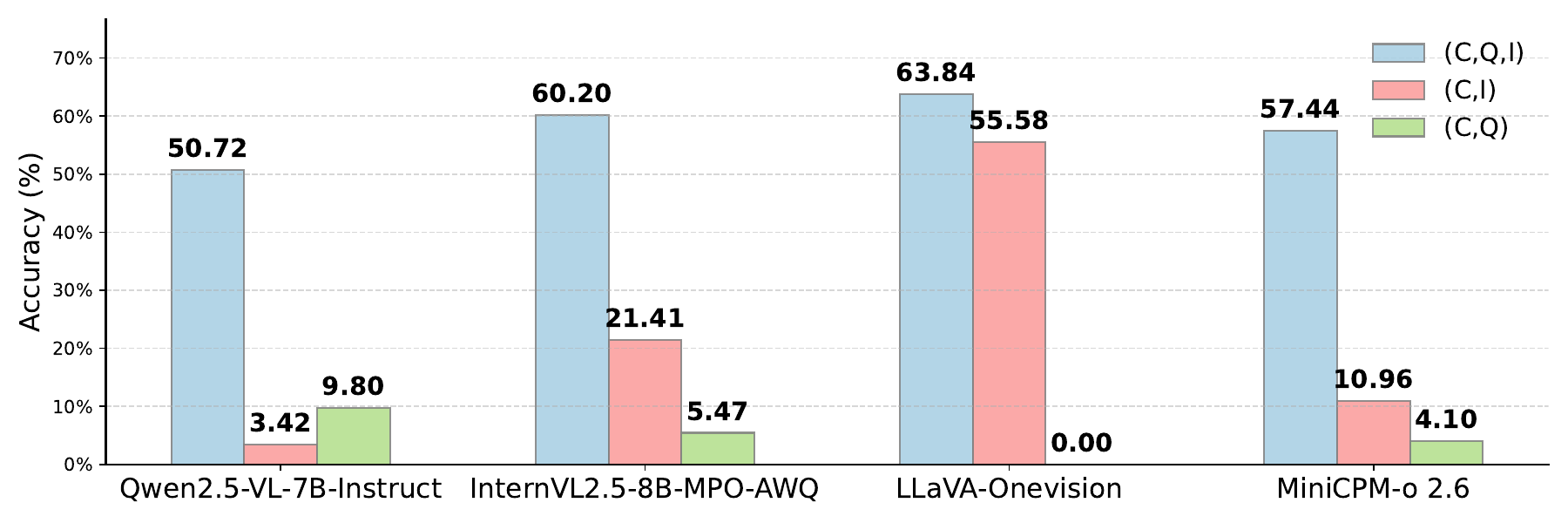}
    \caption{Accuracy under different input settings. $\mathcal{C}$ = Context, $\mathcal{Q}$ = Question, $\mathcal{I}$ = Image. For example, $\mathcal{(C, Q, I)}$ includes all three; $\mathcal{(C, Q)}$ and $\mathcal{(C, I)}$ include only the specified components.}
    \label{fig:accuracy_comparison}
\end{figure}
\section{Analysis and Discussion}

\subsection{Ablation Study}

Figure~\ref{fig:accuracy_comparison} reports the accuracy of four models under three input settings: $\mathcal{(C, Q, I)}$, $\mathcal{(C, I)}$, and $\mathcal{(C, Q)}$. All models achieve the highest accuracy with full input, with LLaVA-OneVision reaching 63.84\%, followed by InternVL2.5 (60.20\%), MiniCPM-o 2.6 (57.44\%), and Qwen2.5 (50.72\%). Removing the question while keeping context and image $\mathcal{(C, I)}$ results in moderate drops—for example, LLaVA drops to 55.58\%, and InternVL2.5 to 21.41\%. In contrast, removing the image $\mathcal{(C, Q)}$ leads to drastic degradation: LLaVA drops to 0.00\%, and Qwen2.5 to 3.42\%. This highlights the essential role of image information in resolving multimodal ambiguity.


\begin{table}[t]
\centering
\small
\resizebox{0.46\textwidth}{!}{
\begin{tabular}{l|cc}
\toprule
\multirow{2}{*}{\textbf{Model}} & \multicolumn{2}{c}{\textbf{Overall}} \\
&\textbf{Acc\textsubscript{\textit{p}}} & \textbf{Acc\textsubscript{\textit{q}}} 
\\\midrule
Qwen2.5-VL-7B-Instruct (w/) & 20.73 & 50.72 \\
Qwen2.5-VL-7B-Instruct (w/o) & 7.06 (\textcolor{red}{↓13.67}) & 12.57 (\textcolor{red}{↓38.15}) \\
InternVL2.5-8B-MPO-AWQ (w/) & 30.98 & 60.20 \\
InternVL2.5-8B-MPO-AWQ (w/o) & 4.56 (\textcolor{red}{↓26.42}) & 8.38 (\textcolor{red}{↓51.82}) \\
LLaVA-OneVision (w/) & 34.62 & 63.84 \\
LLaVA-OneVision (w/o) & 0.00 (\textcolor{red}{↓34.62}) & 0.66 (\textcolor{red}{↓63.18}) \\
MiniCPM-o 2.6 (w/) & 26.42 & 57.44 \\
MiniCPM-o 2.6 (w/o) & 1.14 (\textcolor{red}{↓25.28}) & 4.63 (\textcolor{red}{↓52.81}) \\
\bottomrule
\end{tabular}}
\caption{Ablation study: Only input $\mathcal{(Q, I)}$, with confusion options. Performance drop (\textcolor{red}{↓}) indicates the gap compared to full input.}
\label{tab:results5}
\end{table}
Table~\ref{tab:results5} presents the ablation results using only question and image inputs, with confusing options included. All models exhibit significant performance degradation when context is removed. For example, LLaVA-OneVision drops from 63.84\% to 0.66\% in Acc\textsubscript{q} (↓63.18), and from 34.62\% to 0.00\% in Acc\textsubscript{p} (↓34.62). Similar trends are observed for InternVL2.5 and MiniCPM-o, which also suffer large drops in both metrics. These results underscore the importance of contextual information in resolving ambiguity, especially in the presence of visually or semantically confusing alternatives.
\subsection{Discussion}
As shown in Figure~\ref{fig:casestudy}, this example illustrates how the interpretation of the phrase \textit{``666''} is highly dependent on cultural and visual context, highlighting the necessity of cross-modal disambiguation. \textbf{Scenario 1:} The accompanying image shows the Forbidden City in Beijing, indicating a modern Chinese cultural context. In this setting, ``666'' is widely used as internet slang to express praise, meaning ``awesome'' or ``skillful.''  
\textit{Answer: Positive}. \textbf{Scenario 2:} The image depicts a European Gothic cathedral—Notre-Dame de Paris—evoking a Western Christian context. Here, ``666'' is traditionally associated with the ``number of the beast'' from the Bible, conveying a negative connotation.  
\textit{Answer: Negative}. 

\subsection{Further Exploration: An Agent-Based Framework for Ambiguity Resolution}

\begin{figure}[t!]
    \centering
    \includegraphics[scale=0.25]{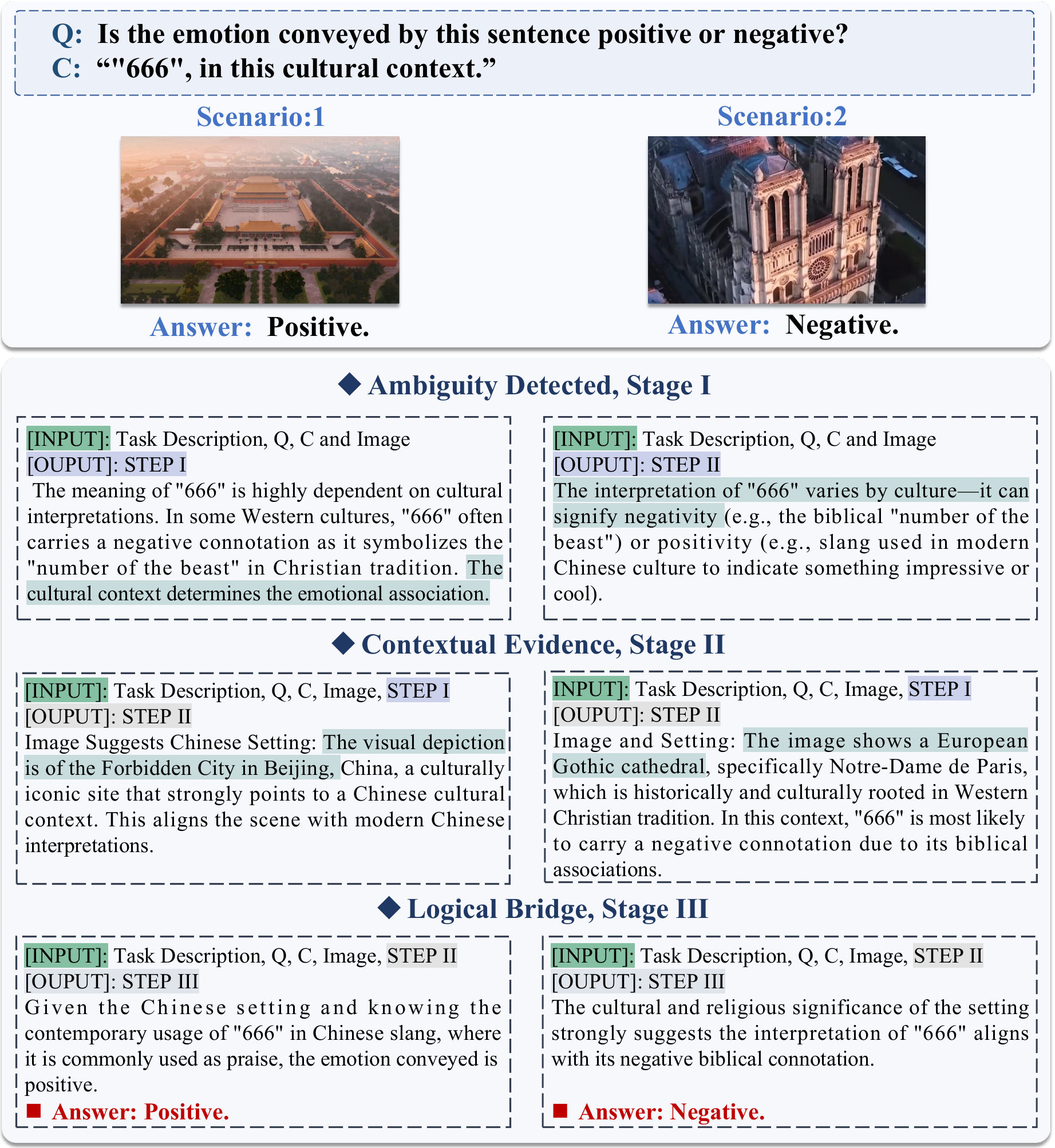}
    \caption{Pipeline of our proposed framework. We first identify ambiguity and its significance from the task description, question, and context/image. Then we generate contextual evidence to resolve the ambiguity. Finally, we bridge the context and image through reasoning to produce the final answer.} 
    \label{fig:casestudy}
\end{figure}

To better address the cross-modal ambiguity resolution, we propose an agent-based framework, as shown in Figure~\ref{fig:casestudy}. First, the model takes the task description, question, and ambiguous context/image as input to identify the ambiguity and its significance. Second, it generates contextual evidence to resolve the ambiguity. Finally, the model bridges the context and image through logical reasoning to produce the final answer.

Formally, given a task description $\mathcal{D}$, a question $\mathcal{Q}$, and a multimodal ambiguous context $\mathcal{X}$ (e.g., an image and text), the agent-based model $\mathcal{M}$ solves the task through a three-step process as illustrated in Figure~\ref{fig:casestudy}:

\noindent \textbf{Step 1: Ambiguity Detection.}  
The model first detects the ambiguity and explains its significance:
\[
\mathcal{A} = \mathcal{M}(\mathcal{D} \oplus \mathcal{Q} \oplus \mathcal{X} \oplus \mathcal{T}_1),
\]
where $\mathcal{T}_1$ is a prompt guiding the model to identify potential ambiguity and why it matters, and $\mathcal{A}$ denotes the ambiguity explanation.

\noindent \textbf{Step 2: Contextual Evidence Extraction.}  
The model then extracts relevant contextual evidence to resolve the ambiguity:
\[
\mathcal{E} = \mathcal{M}(\mathcal{D} \oplus \mathcal{Q} \oplus \mathcal{X} \oplus \mathcal{A} \oplus \mathcal{T}_2),
\]
where $\mathcal{T}_2$ instructs the model to generate explanatory evidence from the context, and $\mathcal{E}$ denotes the extracted evidence.

\noindent \textbf{Step 3: Logical Reasoning and Final Answer.}  
Finally, the model uses evidence to logically align context and image, and generate the final answer:
\[
\mathcal{R} = \mathcal{M}(\mathcal{D} \oplus \mathcal{Q} \oplus \mathcal{X} \oplus \mathcal{E} \oplus \mathcal{T}_3),
\]
where $\mathcal{T}_3$ prompts the model to conduct reasoning and provide the answer $\mathcal{R}$.

\noindent \textbf{Unified Prompt}  
Alternatively, the entire process can be completed with a unified prompt:
\[
\mathcal{A}, \mathcal{E}, \mathcal{R} = \mathcal{M}(\mathcal{D} \oplus \mathcal{Q} \oplus \mathcal{X} \oplus \mathcal{T}),
\]
where $\mathcal{T} = \mathcal{T}_1 \oplus \mathcal{T}_2 \oplus \mathcal{T}_3$.

\begin{table}[t]
\centering
\small
\resizebox{0.49\textwidth}{!}{
\begin{tabular}{l|ccccc|cc}
\toprule
\multirow{2}{*}{\textbf{Model}} & \multicolumn{2}{c}{\textbf{CODIS}} & \multicolumn{1}{c}{\textbf{MMA}} & \multicolumn{2}{c|}{\textbf{MUCAR}} & \multicolumn{2}{c}{\textbf{Overall}}  \\
& \textbf{Acc\textsubscript{\textit{p}}} & \textbf{Acc\textsubscript{\textit{q}}} & \textbf{Acc} & \textbf{Acc\textsubscript{\textit{p}}} & \textbf{Acc\textsubscript{\textit{q}}} & \textbf{Acc\textsubscript{\textit{p}}} & \textbf{Acc\textsubscript{\textit{q}}} \\\midrule\noalign{\vskip -0pt}
Vanilla &  36.26 & 59.49 & 72.0&32.35&53.91&34.31&54.61\\
CoT     &36.81 & 60.76 & 68.0&28.57&56.41&32.69&52.44\\
CODIS    & 36.80 & 60.30 & 71.0&28.29&56.96&32.55&53.19	\\
OURS  & \textbf{42.49} & \textbf{63.46}& \textbf{84.0} &\textbf{44.87}&\textbf{66.78}&\textbf{43.68}&\textbf{64.11}\\
\bottomrule
\end{tabular}}
\caption{Results of our method on CODIS, MMA
 and MUCAR based on \texttt{GPT-4o-2024-11-20.} }
\label{tab:results3}
\end{table}

To demonstrate the generality of our framework across different types of ambiguity, we also report performance of our framework on other wide discussed benchmarks, including CODIS~\citep{codis} and MMA~\citep{wang2024mma}. As shown in Table~\ref{tab:results3}, our method consistently outperforms all baselines across benchmarks. 
Compared to CoT and CODIS-specific prompting, our approach is not only simple but also effective, demonstrating strong potential across different types of ambiguity.

\section{Conclusion}
We present MUCAR, a benchmark designed to evaluate MLLMs in resolving ambiguities across visual, textual, and multilingual contexts. Unlike prior benchmarks, MUCAR targets cross-modal disambiguation through multilingual text and image-text ambiguity cases. Evaluation of 19 state-of-the-art MLLMs reveals a clear gap from human-level performance, highlighting the need for more context-aware and cross-modally grounded models. We also introduce a simple agent-based framework that improves disambiguation through explicit reasoning. 
MUCAR aims to guide future research toward more robust and interpretable multimodal systems in this direction.

\section*{Limitations}
While MUCAR offers a novel and rigorous benchmark for multilingual cross-modal ambiguity resolution, it has several limitations. It covers only three languages, limiting generalizability to low-resource or typologically diverse languages. The curated examples may not capture the complexity and noise of real-world multimodal data. Its partial reliance on GPT-based evaluation introduces potential biases, and the agent-based reasoning framework, though effective in structured tasks, may struggle with open-ended scenarios. Future work should address broader linguistic coverage, real-world settings, and more robust reasoning methods.

\section*{Ethics Statement}

We ensured that all images in our dataset comply with ethical and legal standards. A small portion of images were manually synthesized to cover rare scenarios, and 0.5\% were personally collected; both underwent strict quality checks to minimize bias. The remainder were obtained from platforms with free usage rights (e.g., Unsplash, Pexels, Pixabay). All images were carefully reviewed to ensure quality, fairness, and copyright compliance. Given their high quality and small proportion, synthetic images are unlikely to affect evaluation outcomes.

\section*{Acknowledgments}
This work is supported by the National Natural Science Foundation of China (No. 62306161, 62276152, 62236011).

\bibliography{anthology, custom}
\bibliographystyle{acl_natbib}

\newpage
\appendix

\section{Dataset Details}
\subsection{Dataset Distribution}

\begin{table}[htbp]
\centering
\small
\begin{tabular}{l|ccc|c}
\toprule
\textbf{Category} & \textbf{Chinese} & \textbf{English} & \textbf{Malay} & \textbf{Total} \\
\midrule
Polysemy     & 63  & 64  & 52  & 179 \\
Homonymy     & 42  & 176 & 32  & 250 \\
Grammar      & 64  & 72  & 78  & 214 \\
Semantics    & 24  & 18  & 42  & 84  \\
Specialized  & 26  & 60  & 8   & 94  \\
Cultural     & 24  & 66  & 0   & 90  \\
Dual-Amb     & 126 & 126 & 120 & 372 \\
\midrule
\textbf{Overall} & \textbf{364} & \textbf{582} & \textbf{332} & \textbf{1278} \\
\bottomrule
\end{tabular}
\caption{Distribution of ambiguity categories across different languages.}
\label{tab:distribution}
\end{table}

\noindent\textbf{MUCAR} Table~\ref{tab:distribution} summarizes the distribution of seven ambiguity types across \textbf{Chinese}, \textbf{English}, and \textbf{Malay}. \textbf{English} has the most annotated instances (582), followed by \textbf{Chinese} (364) and \textbf{Malay} (332). \textit{Homonymy} is notably dominant in English, while \textit{Dual-Ambiguity} remains consistently high across all languages. \textit{Cultural} ambiguity appears in Chinese and English but is absent in Malay. \textit{Grammar} and \textit{Polysemy} are relatively balanced, whereas \textit{Specialized} and \textit{Semantics} vary more significantly. These trends reflect both shared and language-specific ambiguity patterns.

\begin{table}[ht]
\centering
\small
\begin{tabularx}{\linewidth}{X}  
\toprule
\multicolumn{1}{c}{\textbf{Prompt for GPT-4o evaluation.}} \\
\midrule
I'll give you an image. Please answer my question based on the image. \\
Directly select the correct option (A, B, C, D, or E). \\
Use the following format to answer: \\
\\
Answer: [ONLY the option letter; not a complete sentence] \\
\\
Only give me the reply according to this format, don't give me any other words. \\
Now, please answer this question. \\
\\
Question: [QUESTION HERE] \\
Options: [OPTIONS HERE] \\
\bottomrule
\end{tabularx}
\caption{Prompt for GPT-4o evaluation.}
\label{tab:prompt_GPT-4o_evaluation}
\end{table}

\begin{table}[h]
\centering
\small  
\resizebox{0.45\textwidth}{!}{  
\begin{tabular}{l|c}
\toprule
\textbf{Model} & \textbf{Consistency Score} \\
\midrule
GPT-4o & 96.25 \\
Gemini-2.0-flash & 96.25 \\
Qwen2.5-VL-7B-Instruct & 97.35 \\
InternVL2.5-8B-MPO-AWQ & 97.57 \\
mPLUG-Owl3-7B-240728 & 95.26 \\
\bottomrule
\end{tabular}}
\vspace{-6pt}
\caption{Consistency results of different models.}
\label{tab:consistency_results}
\vspace{-6pt}
\end{table}

\begin{table*}[t]
\centering
\small
\resizebox{0.98\textwidth}{!}{
\begin{tabular}{l|c|c|c|c}
\toprule
\textbf{Name} & \textbf{Gender} & \textbf{Fluent Languages} & \textbf{Country/Region} & \textbf{Professional Background} \\
\midrule
author1 & Male & Chinese, English & China & Computer Science, Telecommunications, Systems Science \\
author2 & Female & English, Chinese & England & Computer Science, Telecommunications, Finance, Law \\
author3 & Female & Malay, English, Chinese & Malaysia & Computer Science \\
author4 & Male & Chinese, English & China & Computer Science \\
author5 & Female & Chinese, English & China & Computer Science, Economics \\
author6 & Female & Chinese, English & China & Computer Science, Economics, Statistics, Data Science, Humanities and Social Sciences \\
author7 & Male & English, Chinese, Japanese & China & Computer Science, Data Science \\
author8 & Male & Chinese, English & China & Computer Science, Statistics \\
\bottomrule
\end{tabular}}
\caption{Information about Annotators}
\label{Information about Annotators}
\end{table*}

\begin{table*}[htbp]
\centering
\resizebox{\textwidth}{!}{ 
\begin{tabular}{lcccccccccccccccc}
\toprule
\textbf{Model} & \multicolumn{2}{c}{\textbf{Polysemy}} & \multicolumn{2}{c}{\textbf{Homonymy}} & \multicolumn{2}{c}{\textbf{Grammar}} & \multicolumn{2}{c}{\textbf{Semantics}} & \multicolumn{2}{c}{\textbf{Specialized}} & \multicolumn{2}{c}{\textbf{Cultural}} & \multicolumn{2}{c}{\textbf{Dual-ambiguity}} & \multicolumn{2}{c}{\textbf{Overall}} \\
 & \textbf{Acc\textsubscript{\textit{p}}} & \textbf{Acc\textsubscript{\textit{q}}} & \textbf{Acc\textsubscript{\textit{p}}} & \textbf{Acc\textsubscript{\textit{q}}} & \textbf{Acc\textsubscript{\textit{p}}} & \textbf{Acc\textsubscript{\textit{q}}} & \textbf{Acc\textsubscript{\textit{p}}} & \textbf{Acc\textsubscript{\textit{q}}} & \textbf{Acc\textsubscript{\textit{p}}} & \textbf{Acc\textsubscript{\textit{q}}} & \textbf{Acc\textsubscript{\textit{p}}} & \textbf{Acc\textsubscript{\textit{q}}} & \textbf{Acc\textsubscript{\textit{p}}} & \textbf{Acc\textsubscript{\textit{q}}} & \textbf{Acc\textsubscript{\textit{p}}} & \textbf{Acc\textsubscript{\textit{q}}} \\
\midrule
Human & 75.00 & 86.59 & 73.50 & 86.00 & 70.09 & 85.05 & 78.57 & 89.29 & 79.55 & 89.36 & 68.29 & 83.53 & 81.43 & 87.12 & 74.66 & 86.42 \\
\bottomrule
\end{tabular}}
\caption{Human Evaluation Results. For humans, the two experimental settings show little difference, so we chose the more challenging \textbf{with confusing options} setting for human evaluation.}
\label{tab:human_evaluation}
\end{table*}

\subsection{Information about Annotators}
In our study, the annotators are the co-authors of the paper. We opted for this arrangement to ensure careful and consistent evaluation, given the nuanced nature of the tasks involving ambiguity and cross-linguistic interpretation. Below, we provide detailed information about the annotators, including their linguistic fluency, country/region of residence, professional background, and gender. Table \ref{Information about Annotators} shows the information about annotators. It is important to note that all the images used in the dataset were carefully selected and manually reviewed. The selection process involved 8 data curators, who cross-checked the images to ensure quality and relevance. Only those images that received unanimous approval from all curators were included in the dataset. This approach ensured that the images met the necessary standards for consistency and quality.

\subsection{Potential Risks Analysis}
As for the details of data collection, we estimate that approximately 2.3\% of the images are AI-generated and 0.5\% of the images are taken personally, while the rest are crawled from online sources (e.g., Unsplash, Pexels, Pixabay). 

Regarding the potential impact of synthetic images on the evaluation of multimodal ambiguity resolution, all synthetic images were carefully crafted to closely resemble real-world scenes and passed strict manual quality checks. They were designed to supplement rare or hard-to-collect cases, enhancing benchmark coverage without introducing bias. Given their high quality and small proportion in the dataset, any potential impact on model performance evaluation is expected to be negligible.

\section{Human Evaluation}

\subsection{Results of Human Evaluation}

We have conducted a more comprehensive human evaluation. Specifically, we conducted evaluations with 3 human evaluators for Chinese and English, and 2 human evaluators for Malay. The final evaluation scores were averaged across the evaluators for each language. These new results provided a concrete basis for comparison against the current MLLMs performance.
Table~\ref{tab:human_evaluation} is the full breakdown of human performance across various categories. For humans, the two experimental settings show little difference, so we chose the more challenging with confusing options setting for human evaluation. As shown in the table, our benchmark features higher difficulty and a larger dataset. The evaluation scores of human evaluators are significantly higher than those of large models, which highlights a strong gap between human-level performance and current MLLMs. Specifically, the best overall $\mathrm{Acc_p}$ of current MLLMs is only 37.32, but human $\mathrm{Acc_p}$ reaches 74.66.

\subsection{Consistency between GPT and Human Evaluation}

Due to the diverse and often unpredictable nature of large model outputs, it is difficult to strictly evaluate them based on exact match criteria, as this approach would not always be accurate. Additionally, human evaluation is highly time-consuming and labor-intensive, especially when dealing with large datasets. So, we adopt GPT-based evaluation when assessing model performance.

To ensure the reliability of GPT-based evaluation, we conducted a consistency analysis between human and GPT judgments across multiple models by using prompt in Table \ref{tab:prompt_GPT-4o_evaluation}. The results in Table \ref{tab:consistency_results} show high agreement rates, where the GPT-based binary judgments aligned with human evaluation in over 94\% of the cases, supporting the feasibility of using GPT-based evaluation in place of manual evaluation.

\section{Prompt for Model Inference}
Table~\ref{tab:prompt_model_inference} presents the detailed prompts used during model testing. In the main experiments, we employed English prompts; additionally, we conducted ablation studies using Chinese and Malay prompts to evaluate the impact of different evaluation languages on the experimental results.
The table also lists the prompts used in the three ablation settings: $\mathcal{(Q, I)}$, $\mathcal{(C, Q)}$, and $\mathcal{(C)}$, where $\mathcal{(Q, I)}$ uses the \textit{Question and Image} as input, $\mathcal{(C, Q)}$ uses the \textit{Context and Question}, and $\mathcal{(C)}$ uses only the \textit{Context}.

\begin{CJK*}{UTF8}{gbsn}

\begin{table*}[ht]
\centering
\small
\begin{tabularx}{\linewidth}{m{3cm}<{\centering}m{\dimexpr\textwidth-3cm-4\tabcolsep}}
\toprule
\multicolumn{2}{c}{\textbf{Prompt for Model Inference}} \\\midrule
\makecell{Main experiment \\ In English} &
I'll give you an image. Please answer my question based on the image.
Directly select the correct option (A, B, C, D, or E).
Use the following format to answer:

Answer: [ONLY the option letter; not a complete sentence]

Only give me the reply according to this format, don't give me any other words.
Now, please answer this question.

Question: [QUESTION HERE]

Options: [OPTIONS HERE]
\\\midrule


\makecell{Ablation Study \\ In Chinese} &
我会给你一张图片。请根据图片回答我的问题。
直接选择正确选项 (A, B, C, D, 或 E)。
请使用以下格式回答：

Answer: [仅为选项字母；不是完整句子]

请只按照此格式回复我，不要给出任何其他文字。
现在，请回答这个问题。

问题：[QUESTION HERE]

选项：[OPTIONS HERE]
\\\midrule

\makecell{Ablation Study \\ In Malay} &
Saya akan berikan anda imej. Sila jawab soalan saya berdasarkan imej tersebut.
Pilih terus pilihan yang betul (A, B, C, D, atau E).
Gunakan format berikut untuk menjawab:

Answer: [HANYA huruf pilihan; bukan ayat penuh]

Berikan saya jawapan mengikut format ini sahaja, jangan berikan perkataan lain.
Sekarang, sila jawab soalan ini.

Soalan: [QUESTION HERE]

Pilihan: [OPTIONS HERE]
\\\midrule

\makecell{Ablation Study \\ Only input $\mathcal{ (Q, I) }$ } &
I'll give you an image. Please answer my question based on the image.
Directly select the correct option (A, B, C, D, or E).
Use the following format to answer:

Answer: [ONLY the option letter; not a complete sentence]

Only give me the reply according to this format, don't give me any other words.
Now, please answer this question.

Question: [QUESTION HERE]

Options: [OPTIONS HERE]
\\\midrule

\makecell{Ablation Study \\ Only input $\mathcal{(C, Q)}$ } &
Please answer my question.
Directly select the correct option (A, B, C, D, or E).
Use the following format to answer:

Answer: [ONLY the option letter; not a complete sentence]

Only give me the reply according to this format, don't give me any other words.
Now, please answer this question.

Question: [QUESTION HERE]

Options: [OPTIONS HERE]
\\\midrule

\makecell{Ablation Study \\ Is $\mathcal{ C }$ ambiguous?} &
Please determine whether this sentence is ambiguous. 
If the sentence is ambiguous, please answer 'Yes.'; otherwise, answer 'No.'. 

Please respond directly with 'yes' or 'no', without any additional content. 

Sentence: [SENTENCE HERE]
\\\bottomrule

\end{tabularx}
\caption{Prompt for model inference.}
\label{tab:prompt_model_inference}
\end{table*}
\end{CJK*}

\section{Evaluated Models}
\label{app:models}
\begin{table*}[t]
\centering
\small
\resizebox{0.98\textwidth}{!}{
\begin{tabular}{lcccc}
\toprule
\textbf{Model} & \textbf{Parameters} & \textbf{Vision Encoder} & \textbf{LLM Backbone} & \textbf{V2L Adapter} \\\midrule

GPT-4V~\citep{openai2023gpt4vision} & \multirow{4}{*}{-} & - & - & - \\
GPT-4o~\citep{gpt-4o} & \multirow{2}{*}{-} & - & - & - \\
Gemini~\citep{gemini2023gemini} & & - & - & - \\
Claude-3.5-Sonnet~\citep{Claude35Sonnet} & & - & - & - \\

\midrule

Deepseek-VL-Tiny~\citep{lu2024deepseekvl}  & \multirow{8}{*}{$> 7\mathrm{B}$}  & SigLIP & DeepSeek LLM & MLP \\
Kimi-VL~\citep{kimiteam2025kimivltechnicalreport} & & MoonViT & Moonlight model & MLP \\
Llama-3.2-Vision-11B~\citep{Llama32Vision11B} & & XAttn LLM & Llama 3.1 & XAttn LLM \\
MiniCPM-o 2.6~\citep{yao2024minicpm} & & SigLIP & Qwen2.5-7B & MLP   \\
Idefics3-8B-Llama3~\citep{Idefics3-8B-Llama3_laurençon2024building} & & SigLIP & Llama-3.1-8B-Instruct & XAttn LLM \\
InternVL2-8B~\citep{InternVL2-8B_chen2024expanding} & & InternViT & internLM2.5-7b-chat & MLP \\
InternVL2.5-8B-MPO~\citep{InternVL2.5-8B-MPO_wang2024mpo} & & InternViT-V2.5 & internLM2.5-7b-chat & MLP \\
InternVL2.5-8B-MPO-AWQ~\citep{InternVL2.5-8B-MPO-AWQ_chen2024expanding} & & InternViT-V2.5 & internLM2.5-7b-chat & MLP \\

\midrule

LLaVA-v1.6-vicuna-7b~\citep{LLaVA_liu2024improvedbaselinesvisualinstruction}  & \multirow{7}{*}{$\leq 7\mathrm{B}$}  & CLIP ViT-L & vicuna-7b-v1.5 & MLP \\
LLaVA-v1.6-mistral-7b~\citep{LLaVA_liu2024improvedbaselinesvisualinstruction} & & CLIP ViT-L & Mistral-7B-Instruct-v0.2 & MLP \\
Qwen2.5-VL-3B-Instruct~\citep{qwen2.5-VL} & & ViT & Qwen2.5 LLM & MLP \\
Qwen2.5-VL-7B-Instruct~\citep{qwen2.5-VL} & & ViT & Qwen2.5 LLM & MLP \\
mPLUG-Owl3-7B-240728~\citep{ye2024mplugowl3longimagesequenceunderstanding} & & SigLIP & Qwen2 LLM & Linear \\
mPLUG-Owl3-2B-241014~\citep{ye2024mplugowl3longimagesequenceunderstanding} & & SigLIP & Qwen2 LLM & Linear \\
LLaVA-OneVision ~\citep{li2024llavaonevisioneasyvisualtask} & & SigLIP & Qwen2 LLM & MLP \\
\bottomrule
\end{tabular}
}
\caption{API-based and open-source MLLMs selected for evaluation.}
\label{tab:models}
\end{table*}
We evaluate a total of 19 models covering a range of scales and architectures. Our evaluated proprietary models include GPT-4V~\citep{openai2023gpt4vision}, GPT-4o~\citep{gpt-4o}, Gemini~\citep{gemini2023gemini}, and Claude-3.5-Sonnet~\citep{Claude35Sonnet}. For open-source models, we include Deepseek-VL-Tiny~\citep{lu2024deepseekvl}, Kimi-VL~\citep{kimiteam2025kimivltechnicalreport}, Llama-3.2-Vision-11B~\citep{Llama32Vision11B}, MiniCPM-o 2.6~\citep{yao2024minicpm}, InternVL2-8B~\citep{InternVL2-8B_chen2024expanding}, InternVL2.5-8B-MPO~\citep{InternVL2.5-8B-MPO_wang2024mpo}, InternVL2.5-8B-MPO-AWQ~\citep{InternVL2.5-8B-MPO-AWQ_chen2024expanding}, LLaVA-v1.6-vicuna-7b~\citep{LLaVA_liu2024improvedbaselinesvisualinstruction}, Qwen2.5-VL-3B-Instruct~\citep{qwen2.5-VL}, Qwen2.5-VL-7B-Instruct~\citep{qwen2.5-VL}. Details of these models are listed in Table~\ref{tab:models}.
Table~\ref{tab:models} presents a comprehensive overview of the Multimodal Large Language Models (MLLMs) evaluated in our benchmark. The models are categorized into two groups: API-based models and open-source models. For each model, we list its parameter size category (greater than or less than 7 billion), the vision encoder architecture, the underlying language model (LLM) backbone, and the employed vision-to-language (V2L) adapter. API-based models such as GPT-4V and Gemini do not publicly disclose architectural details, while open-source models span a variety of encoders (e.g., SigLIP, CLIP ViT, InternViT), LLM backbones (e.g., Llama, Qwen, InternLM), and adapter types (e.g., MLP, Linear, XAttn LLM). This table highlights the diversity in architectural design choices across MLLMs.

Table~\ref{Ablation Study: prompt in Chinese, no confusion options} reports results for Chinese and Malay prompts under two settings: with and without confusing options. The four sections present detailed model performance for each language and setting combination.  
\begin{table*}[t]
\centering
\small
\resizebox{0.98\textwidth}{!}{
\begin{tabular}{l|cccccccccccc|cc|}
\toprule
\multirow{2}{*}{\textbf{Model}} & \multicolumn{2}{c}{\textbf{Polysemy}} & \multicolumn{2}{c}{\textbf{Homonymy}} & \multicolumn{2}{c}{\textbf{Grammar}} & \multicolumn{2}{c}{\textbf{Semantics}} & \multicolumn{2}{c}{\textbf{Specialized}} & \multicolumn{2}{c|}{\textbf{Cultural}} & \multicolumn{2}{c|}{\textbf{Overall}} \\
& \textbf{Acc\textsubscript{\textit{p}}} & \textbf{Acc\textsubscript{\textit{q}}} & \textbf{Acc\textsubscript{\textit{p}}} & \textbf{Acc\textsubscript{\textit{q}}} & \textbf{Acc\textsubscript{\textit{p}}} & \textbf{Acc\textsubscript{\textit{q}}} & \textbf{Acc\textsubscript{\textit{p}}} & \textbf{Acc\textsubscript{\textit{q}}} & \textbf{Acc\textsubscript{\textit{p}}} & \textbf{Acc\textsubscript{\textit{q}}} & \textbf{Acc\textsubscript{\textit{p}}} & \textbf{Acc\textsubscript{\textit{q}}} & 
\textbf{Acc\textsubscript{\textit{p}}} & \textbf{Acc\textsubscript{\textit{q}}}
\\\midrule
\rowcolor{gray!20}\multicolumn{15}{c}{\textbf{In Chinese, without confusing options}} \\\midrule
Qwen2.5-VL-7B-Instruct & 28.57 & 56.34 & 28.71 & 56.88 & 17.65 & 49.26 & 19.05 & 38.10 & 25.00 & 55.17 & 17.07 & 50.59 & 25.06 & 53.91   \\
InternVL2.5-8B-MPO-AWQ & 41.07 & 68.44 & 44.55 & 68.35 & 30.88 & 63.24 & 33.33 & 64.29 & 37.50 & 59.77 & 24.39 & 60.00 & 38.04 & 65.82   \\
LLaVA-OneVision & 24.72 & 55.31 & 29.91 & 54.00 & 17.59 & 50.93 & 23.81 & 50.00 & 22.73 & 47.37 & 14.63 & 51.76 & 22.78 & 52.26 \\
MiniCPM-o 2.6 & 20.22 & 55.31& 40.17 & 66.40& 34.25 & 64.49 & 33.33 & 60.71& 31.82 & 61.05& 14.63 & 45.88 & 30.52 & 60.75 \\
\midrule
\rowcolor{gray!20}\multicolumn{15}{c}{\textbf{In Chinese, with confusing options}} \\\midrule
Qwen2.5-VL-7B-Instruct & 12.36 & 42.46 & 17.95 & 50.40 & 20.37 & 51.40 & 21.43 & 44.05 & 25.00 & 51.58 & 12.20 & 41.18 & 17.77 & 47.74   \\
InternVL2.5-8B-MPO-AWQ & 29.21 & 60.89 & 29.91 & 58.40 & 26.85 & 56.54 & 26.19 & 58.33 & 31.82 & 57.89 & 21.95 & 55.29 & 28.25 & 58.10   \\
LLaVA-OneVision & 7.87 & 44.13  & 23.93 & 50.80 & 12.96 & 50.93 & 11.90 & 44.05 & 13.64 & 43.16 & 7.32 & 45.88 & 14.35 & 47.63\\
MiniCPM-o 2.6 & 19.97 & 48.61 & 27.35 & 54.80 & 26.85 & 55.14 & 28.57 & 55.95 & 13.64 & 46.47 &  12.20 & 35.29 & 22.78 & 51.38\\
\midrule
\rowcolor{gray!20}\multicolumn{15}{c}{\textbf{In Malay, without confusing options}} \\\midrule
Qwen2.5-VL-7B-Instruct & 35.71 & 63.13 & 30.69 & 61.01 & 25.00 & 55.88 & 28.57 & 54.76 & 37.50 & 62.07 & 9.76 & 51.76 & 30.30 & 59.98  \\ 
InternVL2.5-8B-MPO-AWQ  & 39.88 & 68.14 & 36.63 & 66.06 & 26.47 & 62.50 & 28.57 & 61.90 & 35.00 & 60.92 & 26.83 & 62.35 & 34.85 & 65.27    \\
LLaVA-OneVision & 20.22 & 20.28 & 25.64 & 56.00 & 17.60 & 49.53 & 26.19 & 53.57 & 15.91 & 46.32 & 19.51 & 50.58 & 20.73 & 51.60\\
MiniCPM-o 2.6 & 33.71 & 52.60 & 41.03 & 66.80 & 37.96 & 64.95 & 28.57 & 55.95 & 27.27 & 62.11 & 14.63 & 48.24 & 33.49 & 62.29\\
\midrule
\rowcolor{gray!20}\multicolumn{15}{c}{\textbf{In Malay, with confusing options}} \\\midrule
Qwen2.5-VL-7B-Instruct  & 14.61 & 45.25 & 18.80 & 50.00 & 24.07 & 50.00 & 21.43 & 46.43 & 25.00 & 51.58 & 9.76 & 36.47 & 19.13 & 47.63  \\
InternVL2.5-8B-MPO-AWQ  & 23.60 & 53.07 & 21.37 & 53.20 & 24.07 & 55.14 & 23.81 & 57.14 & 27.27 & 55.79 & 12.20 & 44.88 & 22.32 & 53.58   \\
LLaVA-OneVision & 15.73 & 46.93 & 27.35 & 56.00 & 15.74 & 51.41 & 21.43 & 52.38 & 1.37 & 40.00 & 12.20 & 51.76 & 18.45 & 50.72\\
MiniCPM-o 2.6 & 22.47& 48.60 & 30.77 & 59.60 & 37.04 & 61.68 & 19.05 & 44.05 & 18.18 & 53.68 & 9.76 & 36.47 & 26.20 & 53.69 \\
\bottomrule
\end{tabular}}
\caption{Ablation study in four settings: Chinese and Malay prompts, each with and without confusing options.}
\label{Ablation Study: prompt in Chinese, no confusion options}
\end{table*}

As shown in Table~\ref{Ablation Study: prompt in Chinese, no confusion options}, models achieve the highest performance without confusing options, with InternVL2.5-8B-MPO-AWQ reaching 65.82 for Chinese prompts and 65.27 for Malay prompts. When confusing options are introduced, overall accuracy drops noticeably: for Chinese prompts, the top score decreases by 7.72 (65.82 → 58.10), while for Malay prompts, the drop is even larger at 11.69 (65.27 → 53.58). This indicates that confusing options substantially increase task difficulty. Across both languages, Chinese prompts perform slightly better than Malay prompts, though the gap remains small (0.55 without confusing options and 4.41 with confusing options). InternVL2.5-8B-MPO-AWQ consistently achieves the best results across all settings.

\section{More Cases}
\begin{figure*}[h!]
    \centering
    \includegraphics[width=1\linewidth]{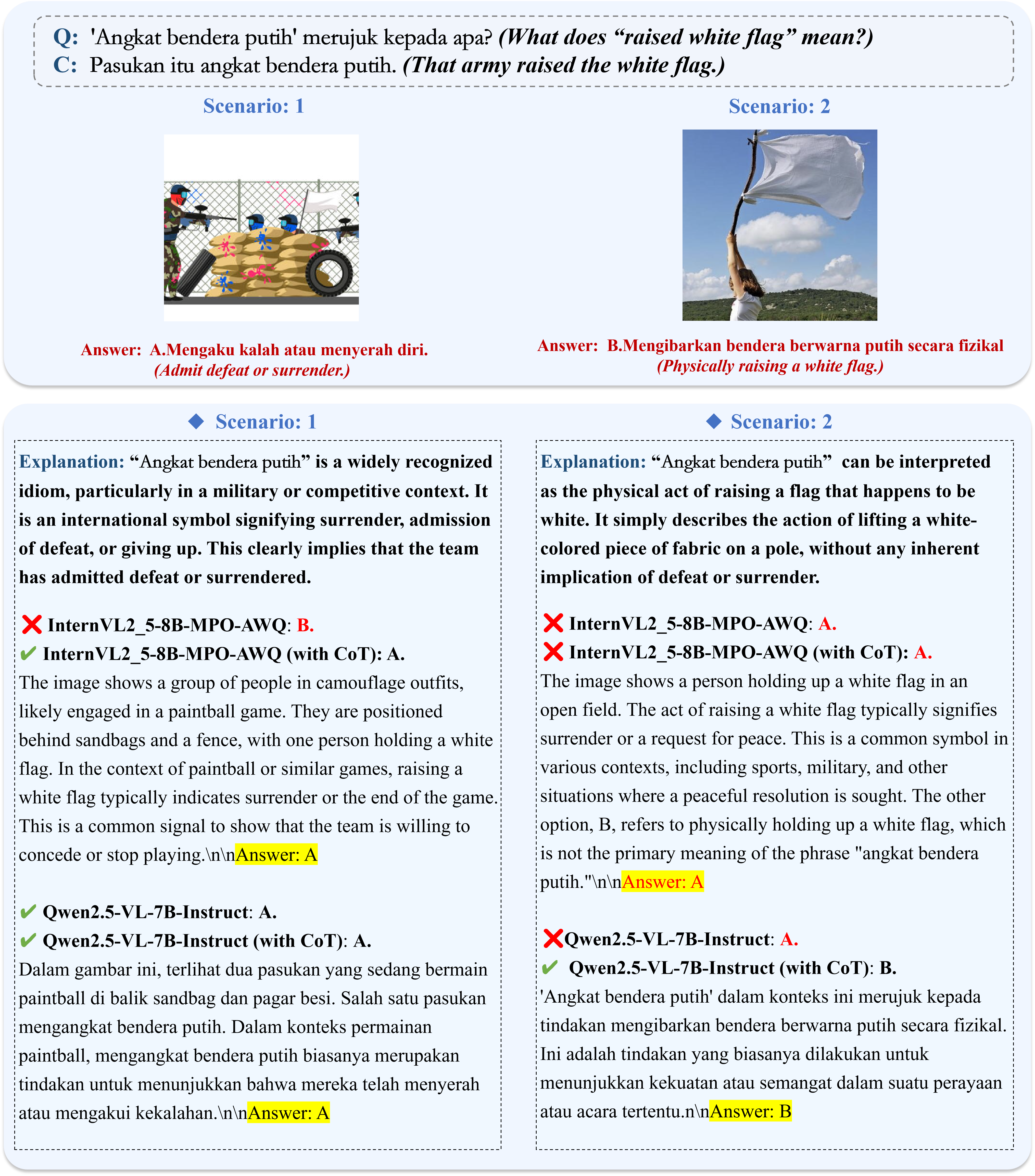}
    \caption{A case of the category of polysemy in Malay.}
    \label{fig:case_study_malay_poly}
    \vspace{-1em}
\end{figure*}

\begin{figure*}[h!]
    \centering
    \includegraphics[scale=0.22]{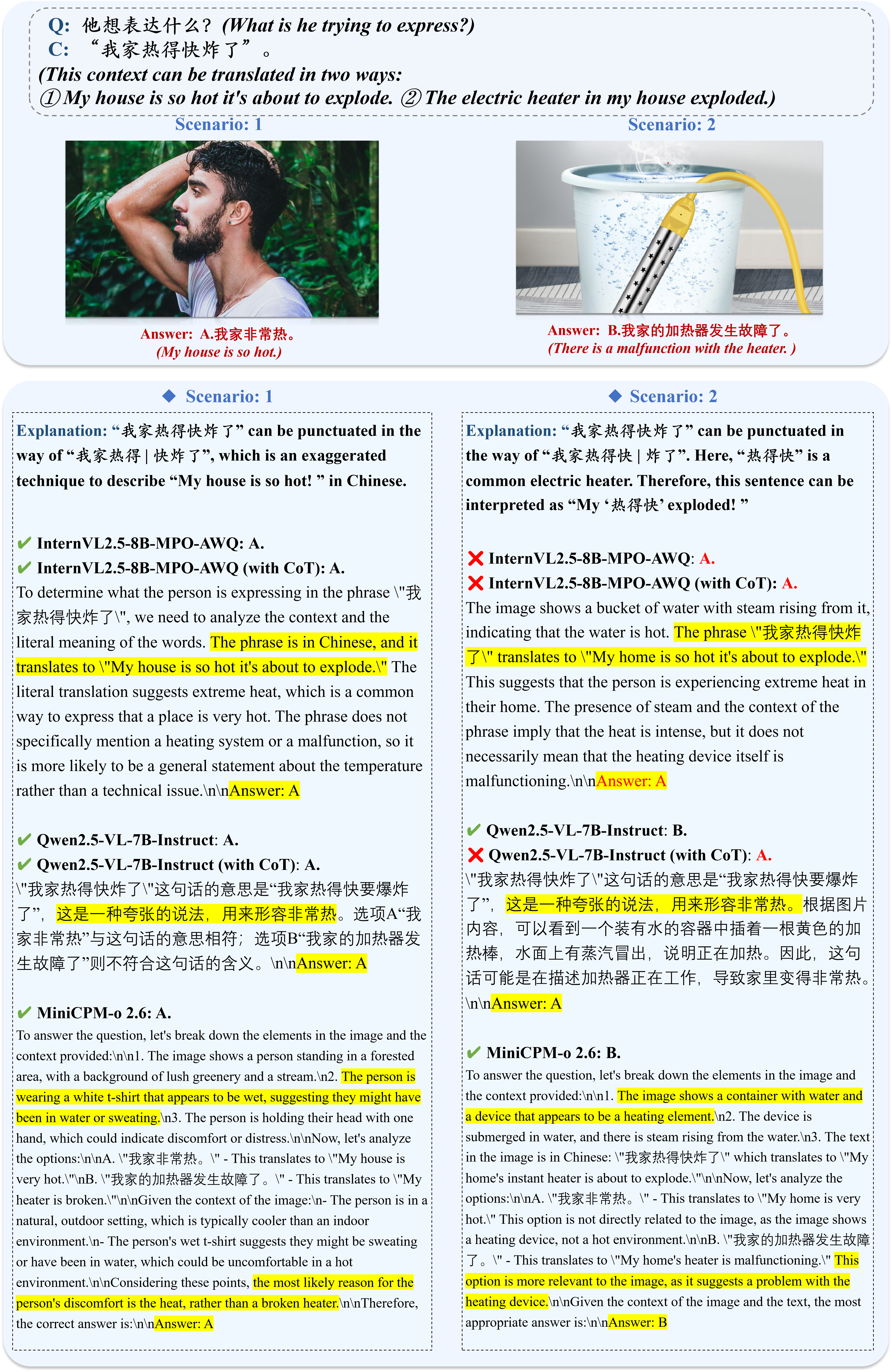}
    \caption{A case of the category of grammar in Chinese.}
    \label{fig:case_study_ReDeKuai}
    \vspace{-1em}
\end{figure*}

\begin{figure*}[h!]
    \centering
    \includegraphics[width=1\linewidth]{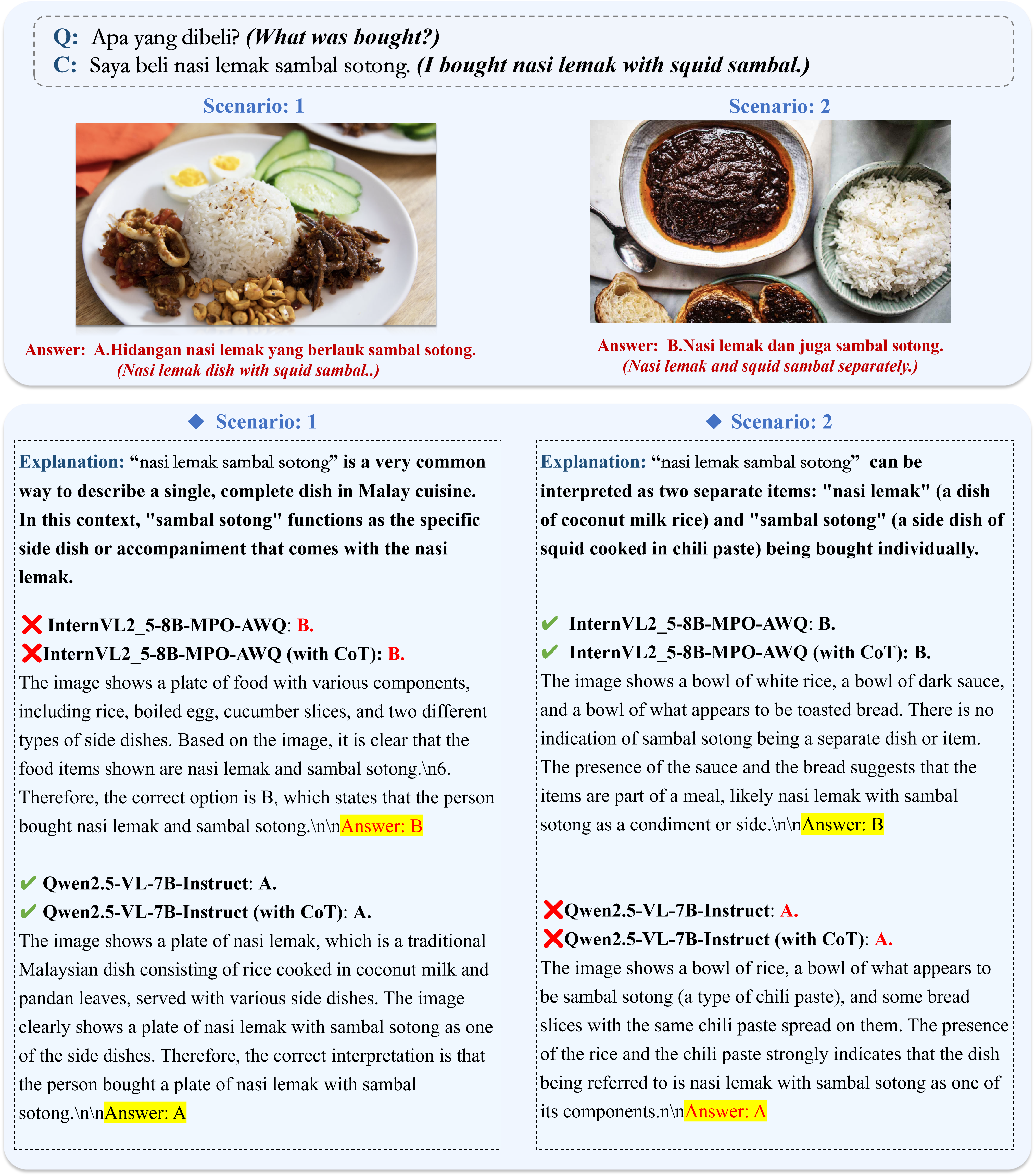}
    \caption{A case of the category of grammar in Malay. }
    \label{fig:case_study_malay_grammar1}
    \vspace{-1em}
\end{figure*}

\begin{figure*}[h!]
    \centering
    \includegraphics[width=1\linewidth]{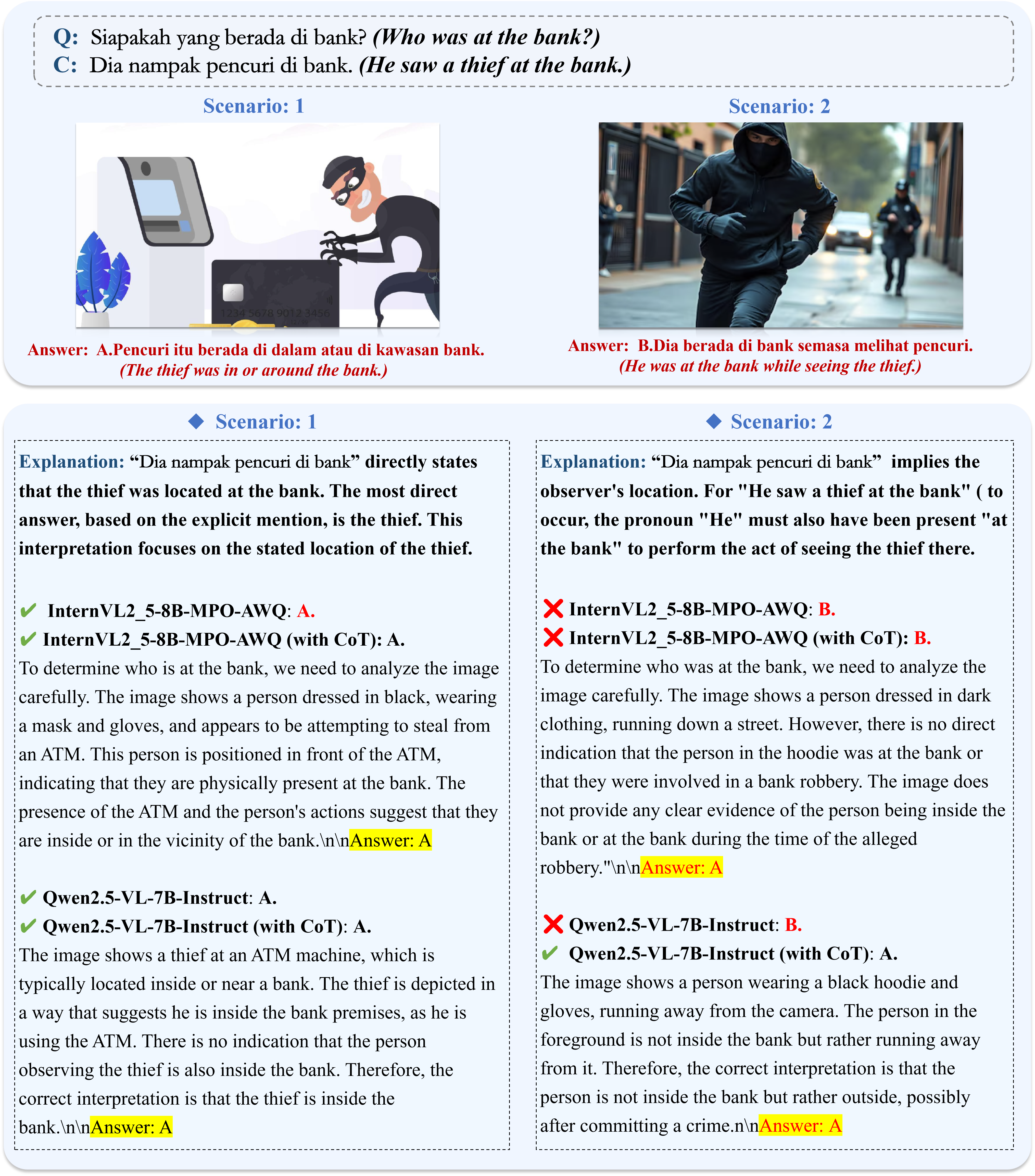}
    \caption{A case of the category of grammar in Malay.}
    \label{fig:case_study_malay_grammar2}
    \vspace{-1em}
\end{figure*}

\begin{figure*}[h!]
    \centering
    \includegraphics[width=1\linewidth]{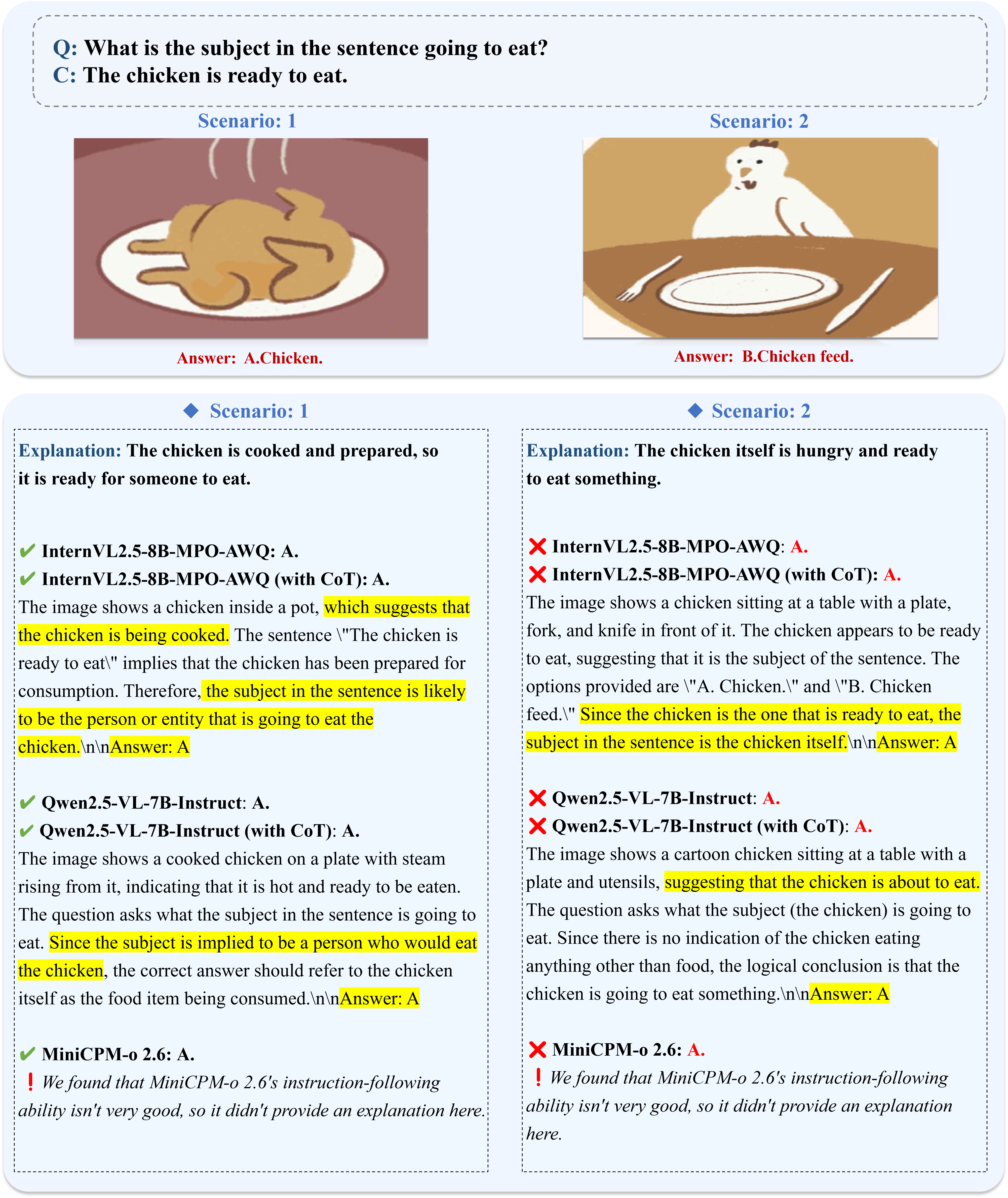}
    \caption{A case of the category of semantics in English.}
    \label{fig:case_study_ChickenEat}
    \vspace{-1em}
\end{figure*}

\begin{figure*}[h!]
    \centering
    \includegraphics[width=0.92\linewidth]{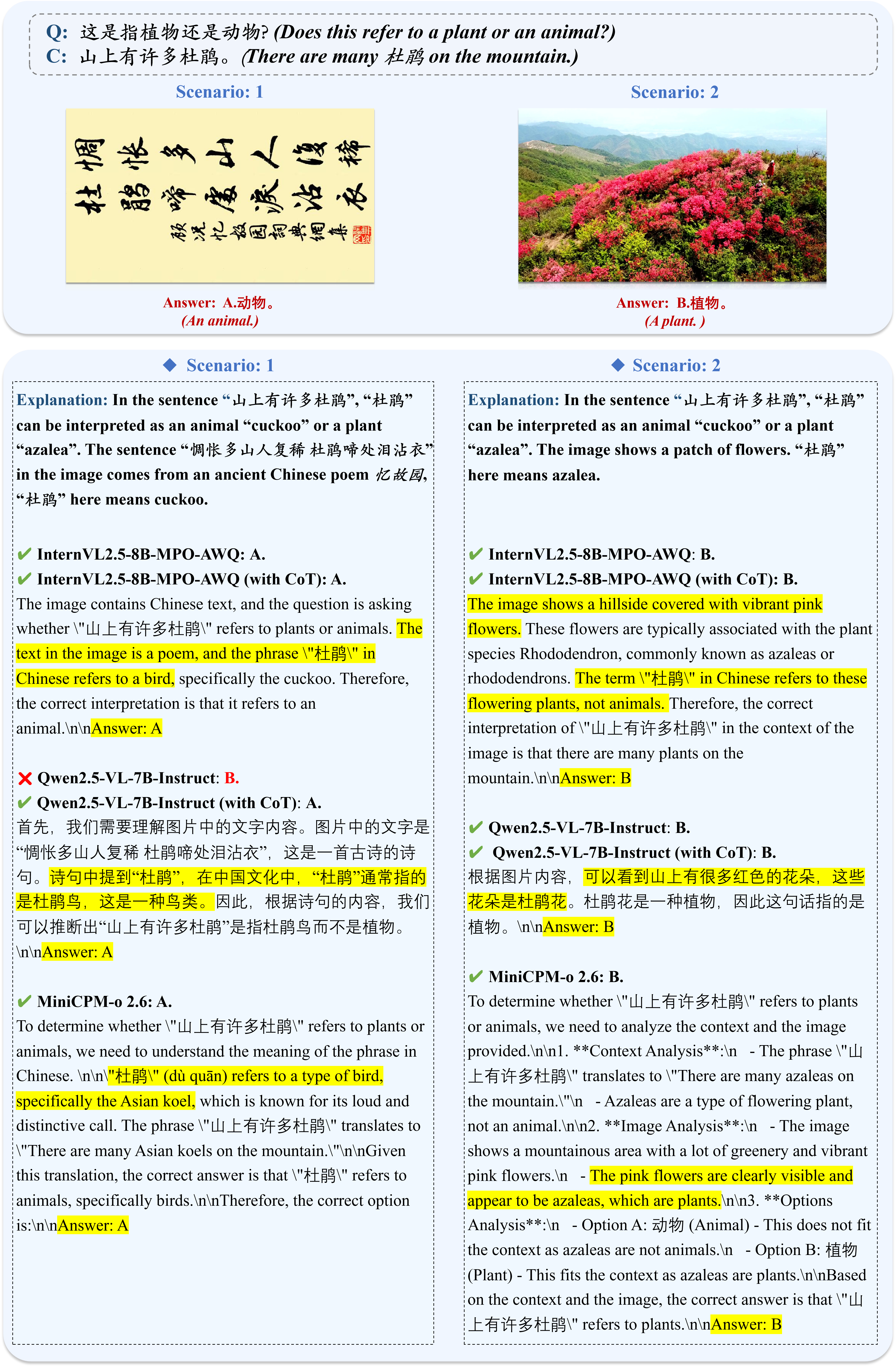}
    \caption{A case of the category of specialized in Chinese.}
    \label{fig:case_study_DuJuan}
    \vspace{-1em}
\end{figure*}

\begin{figure*}[h!]
    \centering
    \includegraphics[width=1\linewidth]{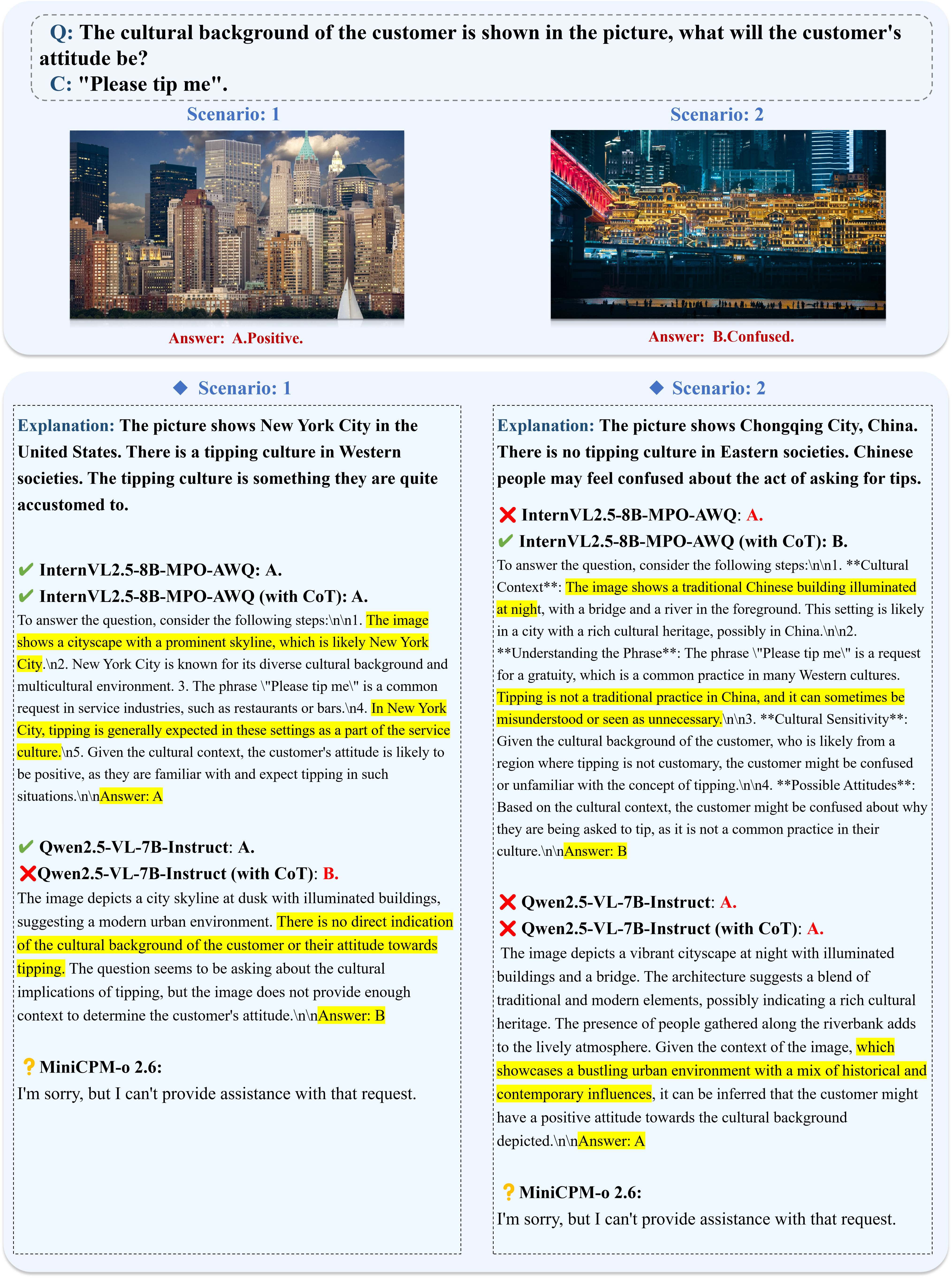}
    \caption{A case of the category of cultural in English.}
    \label{fig:case_study_tipping}
    \vspace{-1em}
\end{figure*}

\begin{figure*}[h!]
    \centering
    \includegraphics[width=1\linewidth]{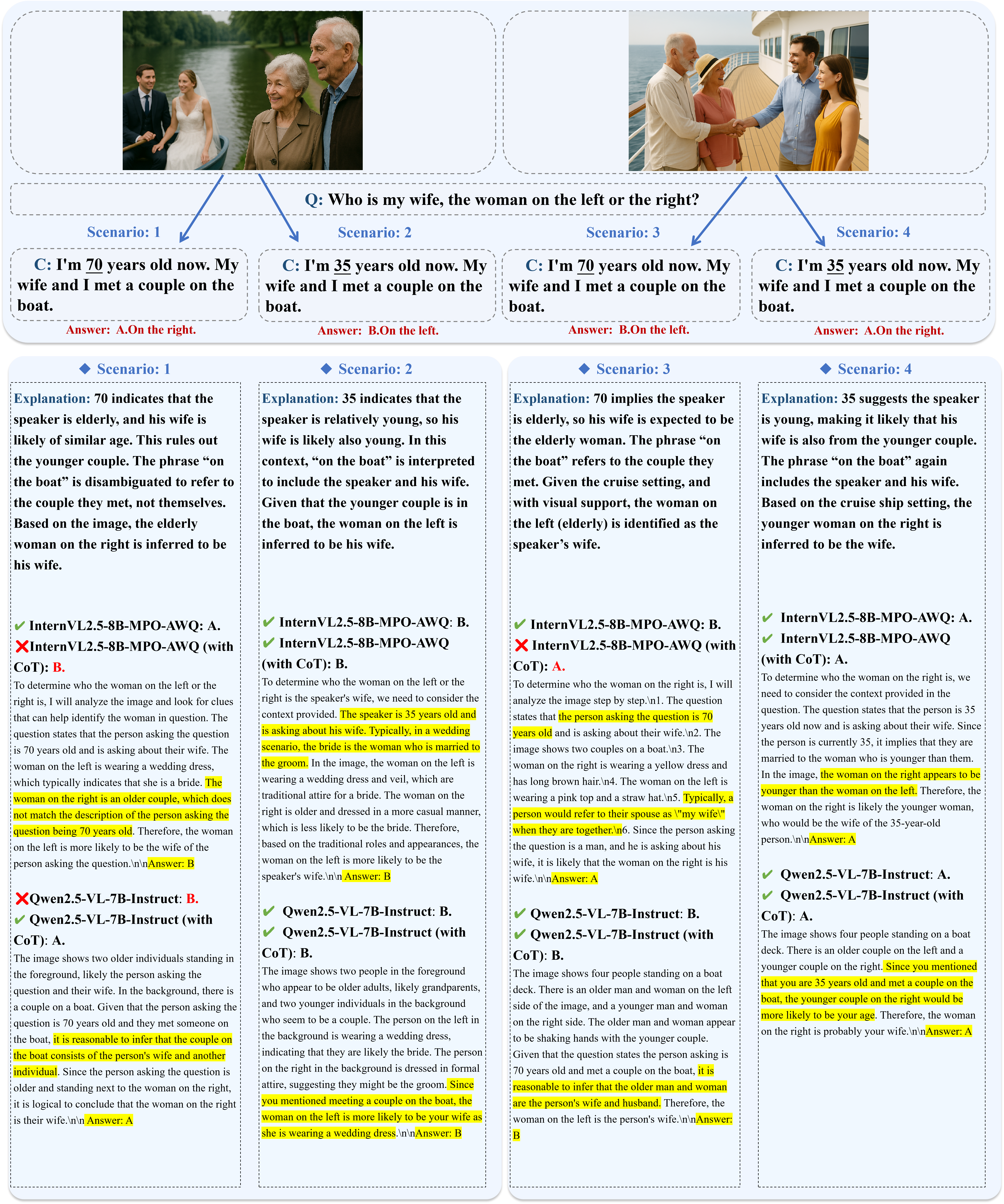}
    \caption{A case of the category of dual-ambiguity in English.}
    \label{fig:case_study_DualAmbiguity}
    \vspace{-1em}
\end{figure*}

We present a comprehensive set of additional cases from the MUCAR dataset to further illustrate the performance of Multimodal Large Language Models (MLLMs). Specifically, Figure~\ref{fig:case_study_malay_poly}, Figure~\ref{fig:case_study_ReDeKuai}, Figure~\ref{fig:case_study_malay_grammar1}, Figure~\ref{fig:case_study_malay_grammar2},  Figure~\ref{fig:case_study_ChickenEat}, Figure~\ref{fig:case_study_DuJuan}, Figure~\ref{fig:case_study_tipping},   Figure~\ref{fig:case_study_DualAmbiguity} display model outputs across various scenarios. For clarity and balance, we select five representative cases from each of the seven predefined categories. In these visualizations, incorrect responses generated by the models are clearly marked in red to allow easy identification of errors.

To support a deeper understanding of the visual and contextual challenges within each case, we also include detailed explanations that highlight the key ambiguities present in the images. These annotations are intended to help readers recognize why a particular question might be difficult to answer correctly, either for a model or a human.

However, it is crucial to emphasize that these explanatory notes were not accessible to either the MLLMs or the human participants during the question-answering process. Both models and human volunteers provided their responses without the benefit of additional contextual guidance, ensuring a fair and unbiased assessment of performance. This setup allows us to isolate and better evaluate the inherent reasoning and perception capabilities of the models in comparison to human interpretation.

Overall, these additional cases offer further insight into the specific limitations and strengths of MLLMs when dealing with multimodal ambiguity, reinforcing the broader findings of our evaluation.

\end{document}